\documentclass[print,twoside]{ieeecolor}
\usepackage{graphicx}
\usepackage{generic}
\usepackage{cite}
\usepackage{amsmath,amssymb,amsfonts}
\usepackage{algorithmic}
\usepackage{algorithm}
\usepackage{hyperref}
\hypersetup{hidelinks=true}
\usepackage{textcomp}
\usepackage{float} 

\usepackage{pgfplots}
\usepackage{filecontents}
\usepgfplotslibrary{external}
\pgfplotsset{/pgf/number format/read comma as period}

\begin{document}

\title{Path Tracking with Dynamic Control Point Blending for Autonomous Vehicles: An Experimental Study}

\author{Alexandre Lombard, Florent Perronnet, Nicolas Gaud, Abdeljalil Abbas-Turki
\thanks{This manuscript has been submitted to IEEE Transactions on Intelligent Vehicles. This work was supported by ANR/xHUB.}
\thanks{All authors are with UTBM, CIAD, F-90010 Belfort cedex, France.}
\thanks{Corresponding author: alexandre.lombard@utbm.fr}}

\maketitle

\begin{abstract}
This paper presents an experimental study of a path-tracking framework for autonomous vehicles in which the lateral control command is applied to a \emph{dynamic control point} along the wheelbase. Instead of enforcing a fixed reference at either the front or rear axle, the proposed method continuously interpolates between both, enabling smooth adaptation across driving contexts, including low-speed maneuvers and reverse motion. The lateral steering command is obtained by barycentric blending of two complementary controllers: a front-axle Stanley formulation and a rear-axle curvature-based geometric controller, yielding continuous transitions in steering behavior and improved tracking stability. In addition, we introduce a curvature-aware longitudinal control strategy based on virtual track borders and ray-tracing, which converts upcoming geometric constraints into a virtual obstacle distance and regulates speed accordingly. The complete approach is implemented in a unified control stack and validated in simulation and on a real autonomous vehicle equipped with GPS-RTK, radar, odometry, and IMU. The results in closed-loop tracking and backward maneuvers show improved trajectory accuracy, smoother steering profiles, and increased adaptability compared to fixed control-point baselines.
\end{abstract}



\section{Introduction}
\label{sec:introduction}

Autonomous vehicles (AVs) have become a focal point of research and development due to their potential to revolutionize transportation systems, enhance road safety, and reduce traffic congestion \cite{OLAYODE20231037}. Central to the functionality of AVs is the control system, which governs the vehicle's movement and ensures it follows a predefined trajectory with high precision. Among the various control mechanisms, lateral control, which pertains to the vehicle's ability to follow a desired path, remains a critical and challenging aspect \cite{ARTUNEDO2024100910}.

Traditional lateral control strategies primarily focus on adjusting the steering angle to minimize the lateral deviation between the vehicle's actual path and the desired trajectory, in terms of distance to the path and relative angle to the path \cite{dominguez2016comparison}. These strategies typically apply the control command at a fixed point, usually the center of mass or a specific point along the vehicle's length, which influences the vehicle's heading and path-following capabilities. However, this conventional approach can be limiting, especially in scenarios that require high maneuverability or when dealing with diverse road conditions.

\begin{figure}
    \centering
    \includegraphics[width=0.99\linewidth]{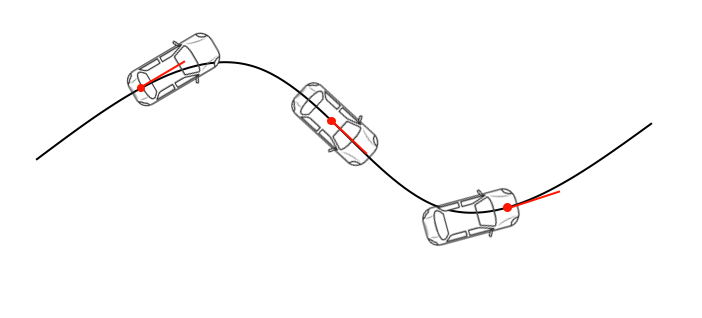}
    \caption{Three different control point for lateral control, and their impact on the behavior of a vehicle during track following}
    \label{fig:control_point}
\end{figure}

In this paper, we introduce a novel lateral control command for autonomous vehicles that offers significant advances over existing methods. The key innovation of our approach lies in its ability to control the point of application of the lateral command. This flexibility allows for the selection of either the front or the rear of the vehicle as the reference point for trajectory following, which impacts the behavior of the vehicle, as illustrated in Fig. \ref{fig:control_point}. In doing so, the proposed control command improves the vehicle's adaptability to various driving scenarios and improves its overall trajectory tracking performance.

\begin{figure}
    \centering
    \includegraphics[width=0.8\linewidth]{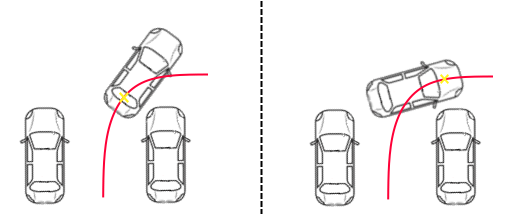}
    \caption{Example: for backward parking, a trajectory is defined according to obstacles, the selection of the control point makes it possible to park}
    \label{fig:control_point_example}
\end{figure}

The ability to switch the point of application of the control command provides several benefits. For instance, controlling the front of the vehicle to follow the trajectory can lead to more precise cornering and better handling in tight spaces, which is particularly advantageous in urban environments. Conversely, controlling the rear of the vehicle can improve stability and safety during high-speed maneuvers or when traversing slippery surfaces. During maneuvers (like reverse parking), being able to dynamically adjust the control point can be mandatory.
Also, Fig. \ref{fig:control_point_example} illustrates how the ability to choose the control point can ease the process of path planning.

In addition, this approach is combined with a curvature-aware longitudinal control solution that is capable of adjusting the vehicle speed according to the radius of curvature of the path to follow. By mixing the proposed lateral and longitudinal control approaches, the accuracy of the path tracking is greatly improved.

After a brief review of the related works, we begin by detailing the mathematical formulation of the proposed lateral control command. An experimental setup is proposed to evaluate the feasibility and performance of the command in terms of accuracy and stability. The results demonstrate that our approach significantly enhances the vehicle's trajectory tracking accuracy and responsiveness under a wide range of operating conditions.

In summary, this paper presents an innovative approach to lateral control in autonomous vehicles, characterized by the novel ability to dynamically select the point of application of the control command, along with a simple yet robust to control the velocity of the vehicle according to the path curvature. This flexibility not only broadens the operational capabilities of autonomous vehicles but also paves the way for more advanced and adaptable control strategies in the future.

\section{Background}
\label{sec:vehicle_model}

When considering lateral control for vehicles, two types of motion are possible: holonomic and non-holonomic. Holonomic vehicles can move independently along any direction and orientation in their configuration space at will, meaning their controllable degrees of freedom match the system’s total degrees of freedom. In contrast, non-holonomic vehicles are subject to non-integrable velocity constraints that reduce their effective maneuverability, preventing certain instantaneous motions and thus making their controllable degrees of freedom fewer than the configuration space dimension. We focus on non-holonomic vehicles because they more accurately represent real-world mobile systems—such as road vehicles—whose motion is inherently constrained by their mechanical design and contact conditions, and thus demand specific motion planning and control strategies. 

Accurate vehicle modeling is crucial to develop effective lateral control algorithms, especially when implementing an adaptive path tracking system with an adjustable application point. This section introduces the kinematic bicycle model used in this study, to capture the essential aspects of vehicle behavior during lateral maneuvers.

The kinematic bicycle model \cite{polack2017kinematic} is a simplified representation that describes the geometric relationship of a vehicle's motion without considering the dynamic forces acting upon it. This model is particularly useful for low-speed scenarios where tire forces and inertial effects are negligible.

\begin{figure}[H]
    \centering
    \includegraphics[width=0.95\linewidth]{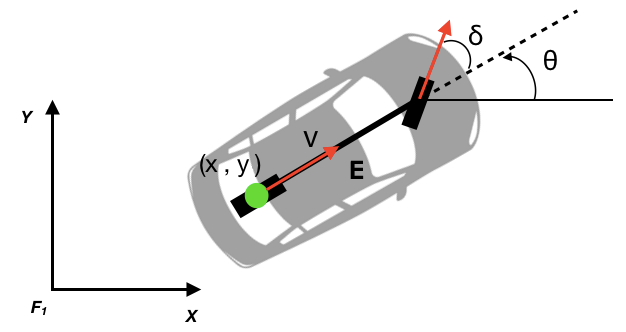}
    \caption{Kinematic bicycle model}
    \label{fig:kinematic_model}
\end{figure}

The state variables are the vehicle's position $(x,y)$, orientation $\theta$, and steering angle $\delta$. The equations of motion are \eqref{eq:kinematic}.

\begin{equation}
    \begin{split}
    \dot x = v \cos(\theta)\\
    \dot y = v \sin(\theta)\\
    \dot \theta = \frac{v}{E} \tan(\delta)
    \end{split}
    \label{eq:kinematic}
\end{equation}

Where: $v$ is the longitudinal velocity, $E$ is the wheelbase (distance between the front and rear axles), as illustrated in Fig. \ref{fig:kinematic_model}.

This model assumes that all wheels are combined into a single front and rear wheel (bicycle approximation), and that the vehicle's lateral motion is solely due to the steering angle $\delta$, which is an acceptable approximation if the vehicle is driven safely \cite{grip2009vehicle}.

\section{Related Work}
\label{sec:related_works}

\subsection{Lateral Control}
\label{sec:lateral_control}

The field of lateral control for autonomous vehicles has seen extensive research in recent decades \cite{ARTUNEDO2024100910}, with numerous approaches proposed to improve the accuracy and robustness of path-following capabilities. This section reviews the most relevant studies and methodologies that have contributed to the development of lateral control systems, highlighting the advances and limitations that our novel approach aims to address.

One of the foundational techniques in lateral control is the Pure Pursuit algorithm, which calculates the steering angle $\delta(t)$ required to follow a circular arc going from the middle of the rear axle to a look-ahead point on the desired trajectory. This method, while simple and effective for smooth paths, can struggle with rapid changes in curvature, leading to oscillations and instability. Also, the pure pursuit gives the expected steering angle to make the middle of the rear axle reach the lookahead point. Thus, according to this command, the rear of the vehicle should follow the trajectory, without any guarantee for the front of the vehicle. In this case, the application point of the command is the middle of the rear axle.

Another widely adopted approach is the Stanley method \cite{thrun2006stanley}, which minimizes both the lateral error and the heading error between the vehicle and the desired path (Equation \ref{eq:stanley}). The Stanley controller has shown significant improvements in stability and accuracy over Pure Pursuit, particularly at higher speeds. However, its performance can degrade in scenarios with low-speed maneuvers or sharp turns due to its fixed control point application. In Stanley's command, the lateral error $e_{fa}(t)$ and the angular error $\theta_e(t)$ are computed according to the front of the vehicle, making it the application point of this command ($v_x(t)$ being the speed of the vehicle and $k$ a gain factor).

\begin{equation}
 \delta(t) = \theta_{e}(t) + tan^{-1}\left(\frac{k e_{fa}(t)}{v_{x}(t)}\right)
 \label{eq:stanley}
\end{equation}

Model Predictive Control (MPC) has also been extensively researched for lateral control in autonomous vehicles \cite{Song2023,kebbati2021optimized}. MPC optimizes a cost function over a finite time horizon, considering both vehicle dynamics and constraints to generate control inputs. This approach provides a more comprehensive framework for handling complex trajectories and dynamic environments. However, the computational complexity and real-time implementation challenges of MPC often limit its practical application in fast-changing scenarios. The level of complexity of the model can also make it harder to apply if it requires data that are difficult to acquire accurately.

More recent studies have explored the use of machine learning and data-driven methods to enhance lateral control. For example, deep-reinforcement learning \cite{li2019reinforcement,liu2024reinforcement} has been applied to develop controllers that can adapt to various driving conditions by learning from experience. While these methods show promise in terms of adaptability and performance, they require extensive training data and can suffer from generalization issues when faced with novel situations. Ensuring their safety and robustness is also a challenge, making them usually unsuitable for real-world applications.

In \cite{lombard2020curvature}, a geometric approach is introduced, described by \eqref{eq:lat}, showing better results than state-of-the-art methods, while being based on geometry, avoiding the issues of data-driven methods.

\begin{equation}
\begin{split}
    \delta_{t} = s \times tan^{-1}\left(\frac{2Ecos(e_{t} - tan^{-1}(\frac{d}{e_{d}}))}{\sqrt{e_{d}^{2} + d^{2}}}\right) + \\sin^{-1}\left(\frac{2E}{L}sin^{-1}\frac{\kappa}{2}\right)
\end{split}
\label{eq:lat}
\end{equation}

Several studies have specifically addressed the point of application of the control command. For example, the research by \cite{falcone2007predictive} proposed a framework for controlling both the front and rear of the vehicle to improve stability during high-speed maneuvers. Similarly, the work by \cite{wang2011improving} introduced a dual-point control strategy that dynamically adjusts the control point based on the driving context. These studies underscore the importance of flexible control point application, but often involve complex control laws and require significant computational resources.

Our work builds upon these foundational studies by introducing a lateral control command that allows for seamless switching between the front and rear application points. This novel approach leverages the strengths of existing methods while addressing their limitations in adaptability and computational efficiency. By providing a more versatile and responsive control mechanism, our method enhances trajectory tracking performance across a wider range of driving scenarios.

In summary, while significant progress has been made in the field of lateral control for autonomous vehicles, challenges remain in achieving high adaptability and precision. Our proposed method contributes to this ongoing research by offering a flexible and efficient solution that dynamically adjusts the point of application of the control command, thereby improving the overall performance and robustness of autonomous vehicle lateral control systems.

\subsection{Longitudinal Control}
\label{sec:longitudinal_control}

Efficient longitudinal control is essential for ensuring vehicle safety and comfort, especially when navigating paths with varying curvature. Adapting the vehicle's speed according to the curvature of the path allows for smoother trajectories, reduces the risk of skidding or loss of control, and enhances passenger comfort. This subsection reviews existing approaches to longitudinal control that consider path curvature, highlighting their methodologies and contributions to the field.

When a vehicle traverses a curved path, the lateral acceleration experienced is proportional to the square of the vehicle's speed and inversely proportional to the radius of curvature 
$a_{lat} = \frac{v^2}{R}$ where  $a_{lat}$ is the lateral acceleration, $v$ is the vehicle's speed, $R$ is the radius of curvature of the path.

Excessive lateral acceleration can lead to passenger discomfort and even vehicle instability. Therefore, adjusting the speed in response to curvature changes is crucial for maintaining safe and comfortable driving conditions.

Model Predictive Control (MPC) has been widely used for longitudinal control due to its ability to handle constraints and predict future states. In curvature-adaptive speed control, MPC algorithms optimize the speed profile by minimizing a cost function that includes terms for tracking error, control effort, and passenger comfort \cite{Guan2024-ot}.

Rule-based systems adjust the vehicle speed based on predefined rules that relate speed to curvature or road conditions. These systems are simpler to implement and require less computational power.

Fuzzy logic controllers handle uncertainties and nonlinearities by using fuzzy sets and inference rules. They are effective in modeling human-like decision-making for speed adaptation.

Reinforcement learning (RL) approaches learn optimal policies through interaction with the environment. In the context of speed adaptation, RL agents aim to maximize a reward function that balances safety, comfort, and efficiency.

\section{Adaptive Path Tracking Algorithm}
\label{sec:adaptive_path_tracking}

\subsection{Adaptive Lateral Control}
\label{sec:adaptive_lateral_control}

The majority of lateral control methodologies are designed to minimize the deviation between a specific reference point on the vehicle and a designated trajectory, in terms of distance and angular difference. This reference point is commonly referred to as the ``control point". Positioning the control point at the front of the vehicle facilitates precise adherence of the front end to the trajectory, whereas locating it at the rear ensures that the back end aligns with the trajectory. Both strategies are viable and are selected based on the specific driving context. For example, placing the control point at the rear is advantageous for reverse maneuvers, such as parking. Consequently, the ability to dynamically adjust the control point enhances the versatility of the vehicle's lateral control, enabling it to adapt more effectively to varying driving conditions.

In this paper, the proposed approach is based on \cite{thrun2006stanley} and \cite{lombard2020curvature}. Both are state of the art lateral control solutions, however, by design \cite{thrun2006stanley} considers the middle of the front axle as the control point, while \cite{lombard2020curvature} considers the middle of the rear axle.

The proposed command is thus expressed by \eqref{eq:adaptive_lc}.

\begin{equation}
    \delta(t) = \alpha \times \delta_{front}(t) + (1 - \alpha) \times \delta_{rear}(t), \alpha \in [0, 1]
    \label{eq:adaptive_lc}
\end{equation}

$\delta_{front}(t)$ can be any command designed to make the front axle follow the track. By application of the method proposed by \cite{thrun2006stanley} expressed by Eq. \eqref{eq:stanley}, $\delta_{front}(t) = \delta_{stanley}(t)$. $\delta_{rear}(t)$ can represent any command designed to make the rear axle follow the track, thus we set $\delta_{rear}(t) = \delta_{cf}(t)$ with $\delta_{cf}$ representing the control output from the method described in \cite{lombard2020curvature}, parameterized for the rear axle. The barycentric weight $\alpha$ corresponds to the relative position of the desired control point with respect to the front and rear axles. This weight governs the blending of the control outputs to achieve the desired control behavior.

Furthermore, in the present paper the command $\delta_{cf}(t)$ is expressed by \eqref{eq:newcf}. 

\begin{equation}
    \begin{split}
        \delta_{cf}(t) = s \times tan^{-1}\left(\frac{2Ecos(e_{t} - tan^{-1}(\frac{d}{e_{d}}))}{\sqrt{e_{d}^{2} + d^{2}}}\right) + \\ tan^{-1}\left(\frac{2E}{L}sin^{-1}\frac{\kappa}{2}\right)
    \end{split}
    \label{eq:newcf}
\end{equation}

With: $s$ equal to 1 if the vehicle is on the left of the trajectory (-1 otherwise), $E$ the wheelbase of the vehicle, $e_{d}$ the signed distance between the rear axle and the trajectory, $e_{t}$ is the angle between the direction of the vehicle and the tangent of the trajectory; $d$ and $L$ are ``lookahead" distances, $\kappa$ is the total curvature change between the direction of the vehicle and the tangent to the trajectory at the lookahead distance $L$. 
Compared to \cite{lombard2020curvature}, the expression has been modified, notably to better control the movement of the back of the vehicle, the parameters are expressed according to the position of the rear axle. The consequences of these changes are detailed in Section \ref{sec:results}.

Also, while being intuitive, the idea of performing a linear interpolation between a command piloting the front axle and another one piloting the rear axle to control any point of the vehicle is difficult to demonstrate theoretically, thus the validation of the command is performed experimentally as described in Section \ref{sec:experimental_setup}.

\subsection{Backward Maneuvers}
\label{sec:backward}

The usage of an adjustable control point enables one to accurately control either the front axle position or the rear axle position. The control of the rear axle position is especially useful during some backward maneuvers; important use cases include reverse parking or docking.

To adapt the control solution to backward maneuvers, we rely on the symmetry of the kinematic bicycle model given by \eqref{eq:kinematic}. Thus, in the case of a reverse movement, the command is expressed by \eqref{eq:newcf_backward}.

\begin{equation}
    \begin{split}
        \delta_{cf,r}(t) = s \times tan^{-1}\left(\frac{2Ecos(e_{t} - tan^{-1}(\frac{d}{e_{d}}))}{\sqrt{e_{d}^{2} + d^{2}}}\right) + \\ tan^{-1}\left(\frac{2E}{L}sin^{-1}\frac{\kappa}{2}\right)
    \end{split}
    \label{eq:newcf_backward}
\end{equation}

\subsection{Virtual Borders Based Longitudinal Control}
\label{sec:virtual_borders}

This section presents a model for adjusting the speed of an autonomous vehicle according to the curvature of the track. 

Indeed, the adaptation of the speed of the vehicle according to the curvature of the track is a key element to ensure the accuracy of the path tracking command. With an excessive speed, the pure-rolling hypothesis (no lateral slip) would not be valid anymore. To ensure we do not fall in this case, a longitudinal control anticipating the curvature is developed.   
The proposed method is inspired by the way that human drivers adjust their speed according to their visibility in corners. It involves creating virtual borders around the track with a parameterized width and employing a ray-tracing technique to generate virtual obstacles ahead. The first intersection between a ray cast from the vehicle and the virtual border serves as a virtual obstacle in a longitudinal control model, which computes the vehicle's acceleration accordingly.

\subsubsection{Virtual Borders}

To incorporate the curvature of the track into the speed adjustment model, virtual borders are established around the track. These borders define the safe driving corridor and are parameterized by a width $w(s)$, which can vary along the path based on road conditions or safety requirements.

Let the centerline of the track be represented by a continuous curve $C(s)$, where $s$ denotes the arc length along the path. The virtual borders $B_{left}$ and $B_{right}$ are defined as offsets from the centerline \eqref{eq:vborders}.

\begin{equation}
\begin{split}
    B_{left}(s) = C(s) + w(s)n(s)\\
    B_{right}(s) = C(s) - w(s)n(s)
\end{split}
\label{eq:vborders}
\end{equation}

Where $n(s)$ is the unit normal vector to the path at point $s$. The parameterized width $w(s)$ allows flexibility in defining the driving corridor, accommodating varying road widths and safety margins. Practically, for a given speed limit, a constant value for $w(s)$ yields satisfying results.

\subsubsection{Ray-Tracing for Virtual Obstacle Detection}

In this method, the ray-tracing method is employed to detect the virtual obstacle ahead of the vehicle. A ray is cast from the vehicle's current position $P_v=(x_v,y_v)$ in the direction of its heading angle $\theta_v$. The ray is mathematically defined by \eqref{eq:ray}.

\begin{equation}
    R(t)=P_v + t \cdot d
    \label{eq:ray}
\end{equation}

Where $t \geq 0$ is a scalar parameter, $d=\left[\cos(\theta_{v}), \sin(\theta_{v})\right]$ is the unit direction vector of the vehicle's heading.

The intersection point $P_{obs}$ between the ray and the virtual borders represents the location of a virtual obstacle. The distance to the virtual obstacle $d_{obs}$ is calculated by \eqref{eq:intersection}.

\begin{equation}
d_{obs}=\min_{t\geq 0}\{ t:R(t)\in B_{left} \cup B_{right}\} 
\label{eq:intersection}
\end{equation}

This distance reflects the available space ahead before the vehicle reaches the virtual boundary, effectively capturing the influence of the track curvature on the vehicle's path.

\subsubsection{Longitudinal Control}

Knowing the distance to the virtual obstacle, any longitudinal control model able to adjust the vehicle speed according to an obstacle can be applied, like Intelligent Driver Model (IDM) \cite{kesting2010enhanced}.

In this paper, we decide to use IDM for \textit{free-flow} acceleration, and to use the command given in \cite{outay2022comparison}, simplified as \eqref{eq:rtacc_stop} to manage obstacles (knowing that the velocity $v_{l}$ of the virtual obstacle is equal to 0), because of its ability to handle the reaction time of the system.

\begin{equation}
    a_{r} = \frac{b_{f}\tau - 2v \pm 2b_{f} \sqrt{\frac{b_{f}b_{l}\tau^2+4b_{l}v\tau - 8b_{l}s}{4b_{f}b_{l}}}}{2\tau}
    \label{eq:rtacc_stop}
\end{equation}

With: $a_r$ the acceleration to apply to the vehicle, $b_{f}$ the comfortable deceleration, $b_{l}$ the emergency deceleration, $s$ the distance to the virtual obstacle, $v$ the speed of the vehicle, and $\tau$ the reaction time of the system.

The resulting longitudinal control is given by taking the minimum acceleration between the free-flow IDM and the equation \eqref{eq:rtacc_stop}, thus $a = \min(a_r, a_{idm})$.

\section{Experimentation}
\label{sec:experimental_setup}

To evaluate the proposed adaptive path-tracking algorithm, simulations as well as real tests. The testing scenarios included two key use cases: trajectory tracking on a straight line and on a circular track. These cases were chosen to demonstrate the stability of the command, and its ability to handle diverse driving conditions, including both continuous path-following and low-speed reverse maneuvers.

The purpose of the tests is to evaluate in a real test-case how the proposed control solution (both longitudinal and lateral) behaves, thus the following are observed:
\begin{itemize}
    \item The relative distance of the vehicle to the track, to evaluate both the accuracy and the stability of the command
    \item The speed of the vehicle, to observe how the turns are managed with the command
    \item The output of the command (\textit{i.e.} the expected steering angle), to evaluate its stability
\end{itemize}

For the tests, 3 cases are considered:
\begin{enumerate}
    \item The control point is placed on the front axle
    \item The control point is placed in the middle position (between the front axle and the rear axle)
    \item The control point is placed on the rear axle
\end{enumerate}
For each case, we test the command forward and backward, totalizing 6 different scenarios.

\subsection{Simulation}

Simulations were conducted using a specifically developed simulator implementing the standard models for the movement of the vehicle. The simulator is available at \url{https://alexandrelombard.github.io/autonomous-driving-sim/}. The command was implemented in a library developed in Rust. 

The measured metrics are the lateral deviation, the heading error, and the control smoothness to assess tracking performance. The presented command is compared with pure-pursuit and Stanley's command for reference (usual parameters are set for these lateral control solutions).

At start the vehicle is placed at 4 meters to the trajectory, then it has to follow a scaled down version of the Nuerbuergring race track, which has several turns of various curvature.

To assess the stability of the command with real sensors, simulations are also performed with simulated uniform noise on the measured data, and a simulated delay between the perception of the environment and the application of the command.

\begin{figure}
    \centering
    \includegraphics[width=0.99\linewidth]{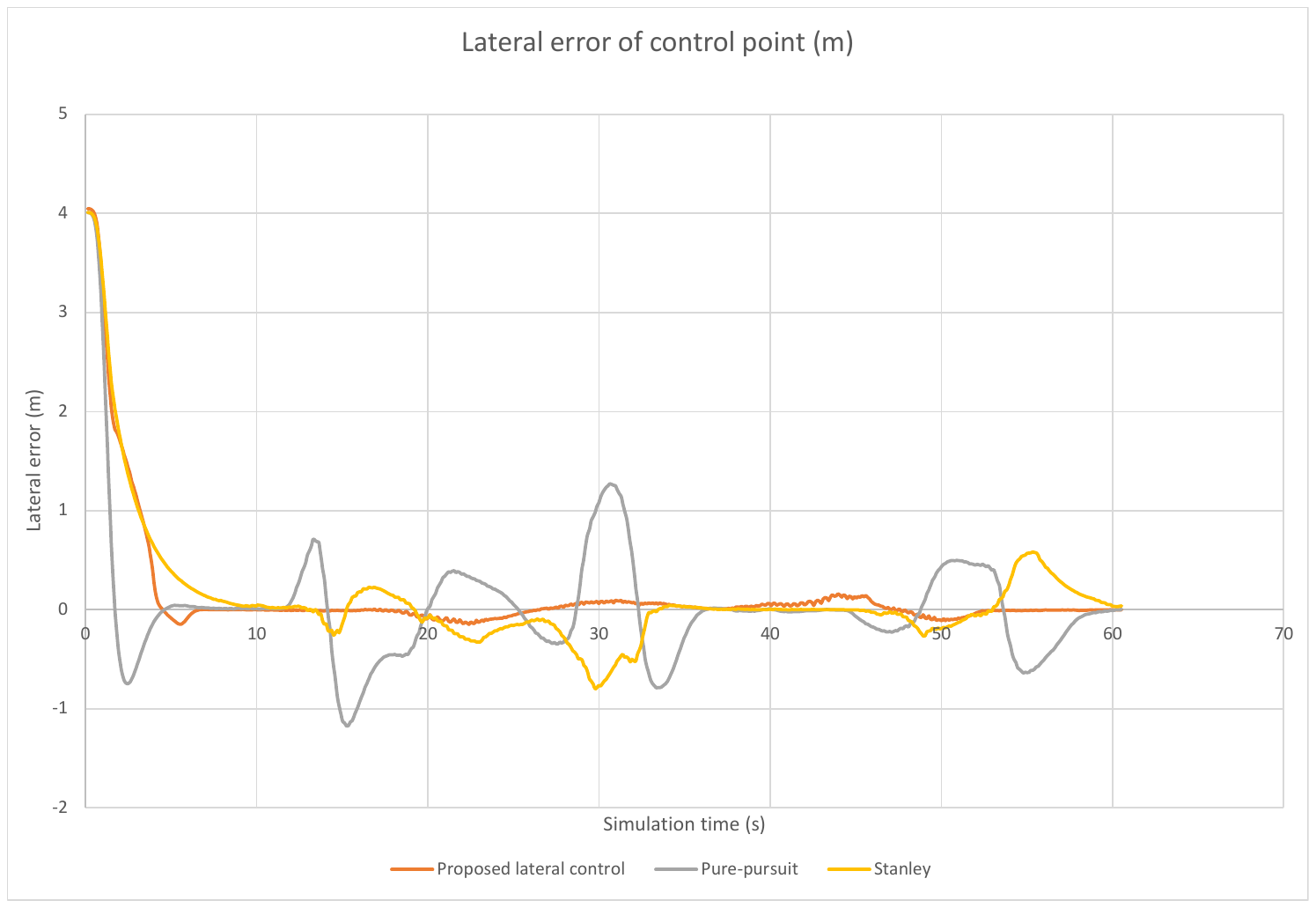}
    \caption{Lateral error of the control point for various command (distance in meters between the control point and its projection on the trajectory)}
    \label{fig:lat_error}
\end{figure}

In Fig. \ref{fig:lat_error}, the lateral error in meters is displayed between the control point and its projection on the track. It can be seen that the lateral error is smaller for the proposed command compared to pure-pursuit and Stanley's command. At the beginning, a small overshoot can be observed, but it is smaller than pure-pursuit, while reaching the trajectory faster than Stanley.

\begin{figure}
    \centering
    \includegraphics[width=0.99\linewidth]{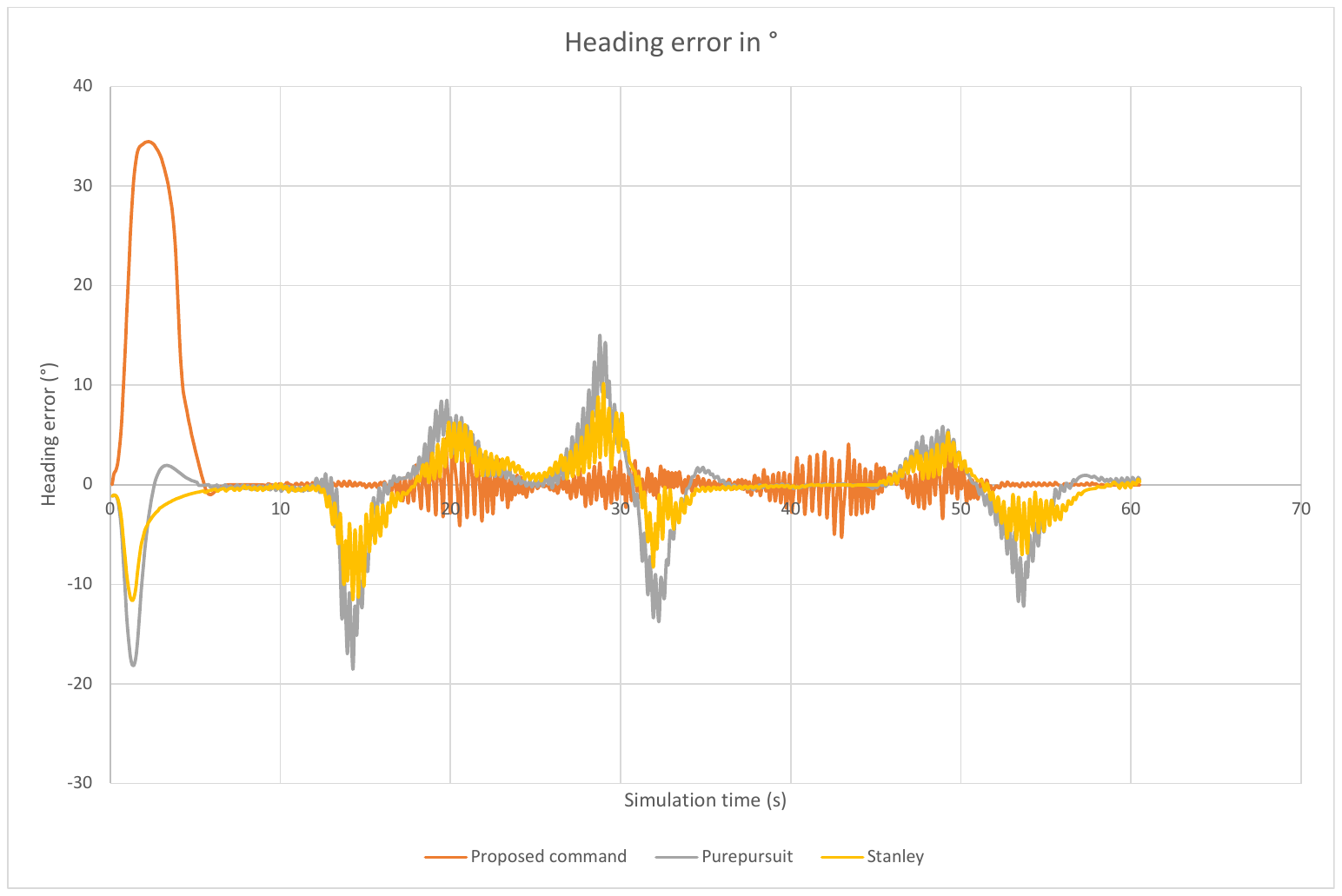}
    \caption{Heading error in degrees for the control point (angle between the heading of the vehicle and the heading of the trajectory at the projection of the control point)}
    \label{fig:heading_error}
\end{figure}

In Fig. \ref{fig:heading_error}, the heading error is displayed. This chart illustrates the stability of the various commands, while the centering of the heading error around 0 of the proposed command shows a better stability than the other two commands. To be noted, the fast oscillations are due to the discretization of the track, leading to a lot of small and quick changes in the heading of the trajectory.

\begin{figure}
    \centering
    \includegraphics[width=0.99\linewidth]{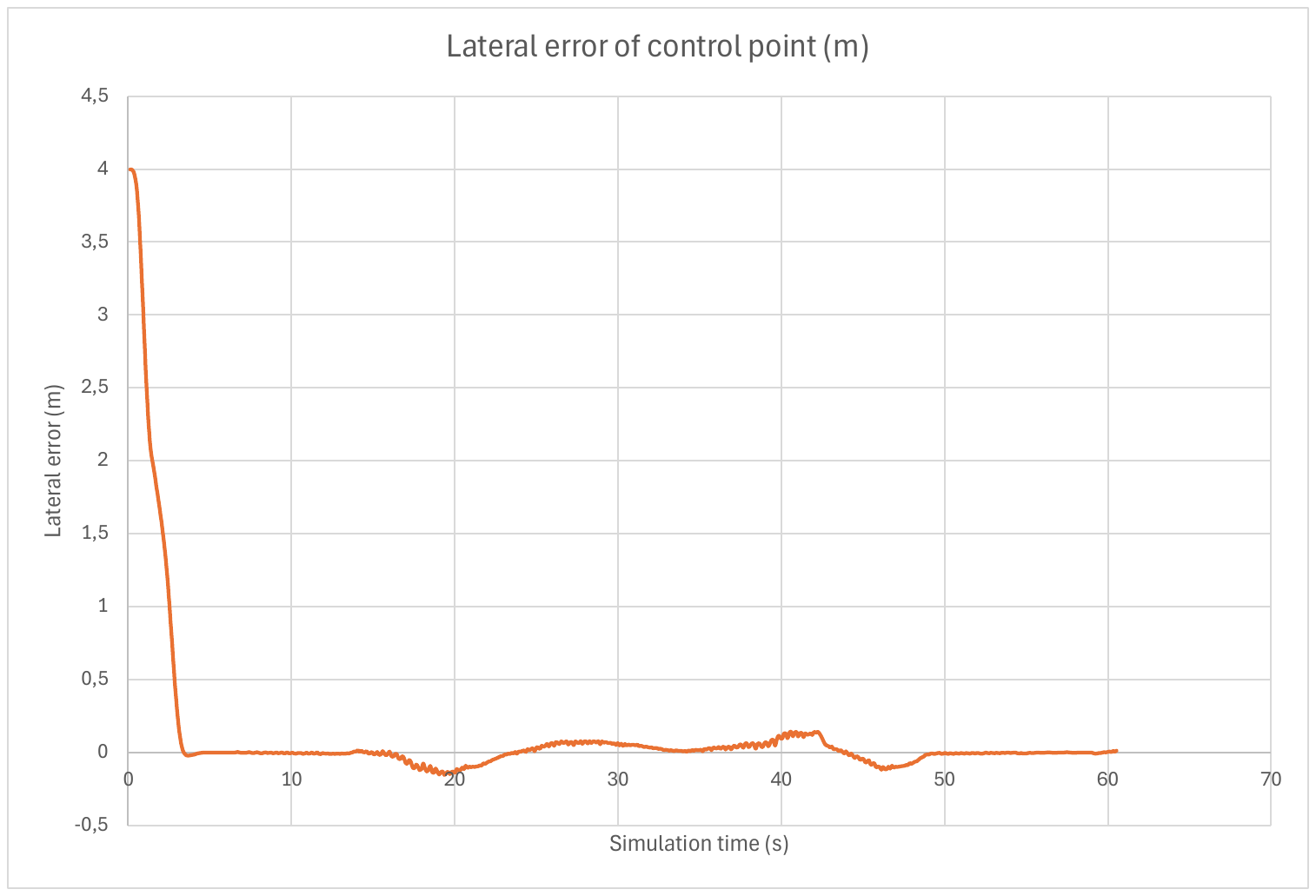}
    \caption{Evolution of the lateral error of the control point using the proposed command and with moving control point}
    \label{fig:varying_cp}
\end{figure}

Finally, the Fig. \ref{fig:varying_cp} illustrates the stability of the command with a dynamic control point thanks to a specific scenario, in which the control point is dynamically moving from the rear of the vehicle to the front. The relative coordinate of the control point is 0 for the rear of the vehicle and 1 for the front, the effective control point is then given by $A(t) = 0.5 \times (sin(\omega t) + 1.0)$ (in this example, $\omega = 0.2$). It can be observed that even if the command alternately tries to stick the front and rear of the vehicle to the track, it is accurate and stable.

\subsection{Real World Validation}

The same Rust library utilized in the simulations was deployed on a real autonomous vehicle to validate the algorithm's effectiveness in real-world scenarios. This direct implementation on the vehicle ensured consistency between the simulation environment and real-world testing, allowing for a comprehensive evaluation of the algorithm's performance under practical constraints.

The unmanned vehicle was equipped with the following sensors:
\begin{itemize}
    \item GPS RTK with an antenna placed close to the center of the wheelbase of the vehicle
    \item Radar
    \item Odometer
    \item Inertial Measurement Unit (IMU)
\end{itemize}

\begin{figure}
    \centering
    \includegraphics[width=0.99\linewidth]{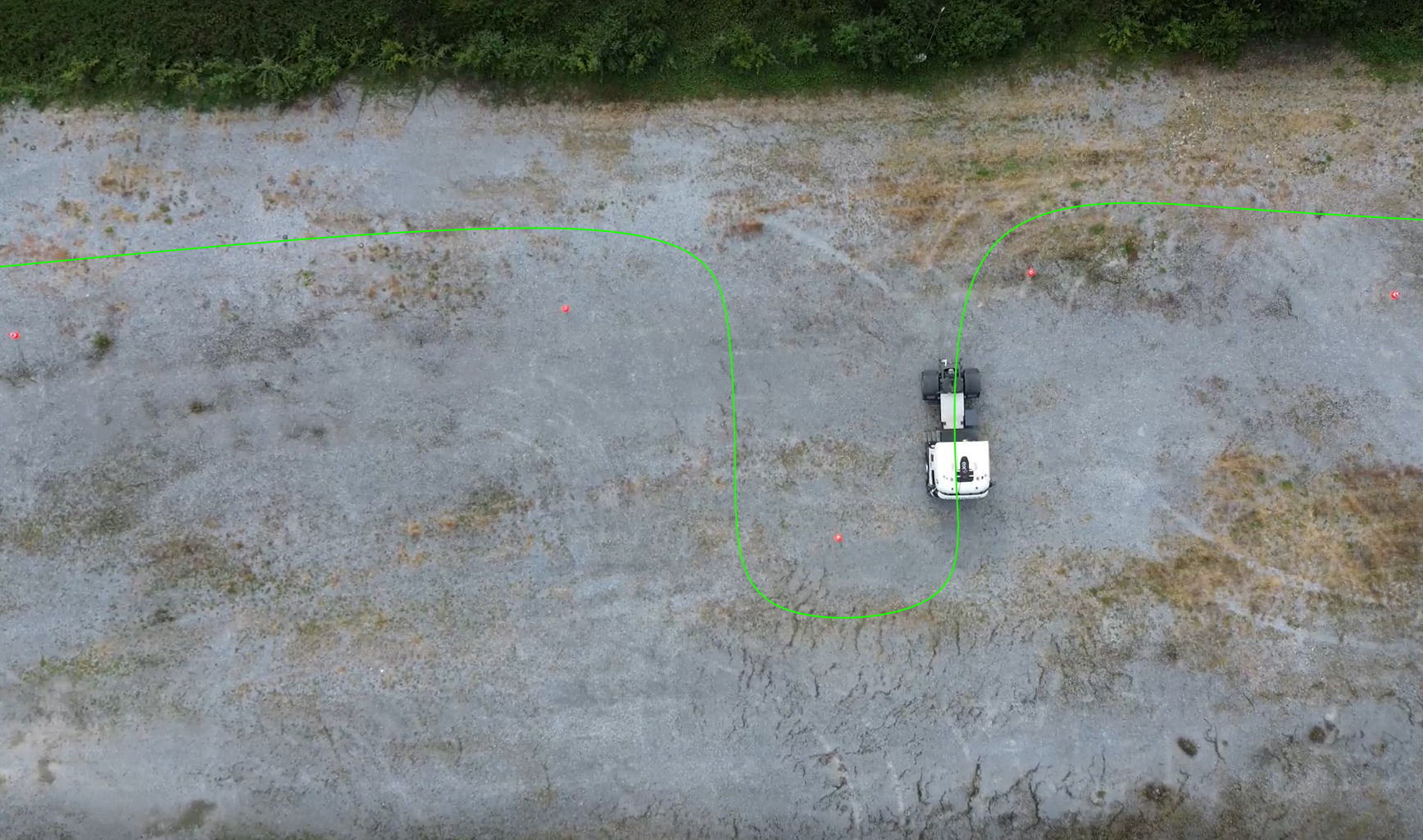}
    \caption{Aerial view of the track, highlighting the section with sudden changes of curvature}
    \label{fig:real_test_track}
\end{figure}

6 tests were performed on a closed loop track, mostly to evaluate the impact of the change of the control point over the trajectory of the vehicle (3 tests forward with the control point on the rear axle, in the middle, and on the front axle, and 3 tests backward). At the beginning, the vehicle is placed at a fixed distance of the track to observe how the control tries to rejoin the track. The parameters applied for the command are given in Table \ref{tab:real_parameters}. The preferred speed has been set to low values as it was dangerous to go faster on the track (especially because of numerous bumps and holes). The Fig. \ref{fig:real_test_track} shows an aerial view of the test track and the vehicle used for the tests.

\begin{table}[]
    \centering
    \begin{tabular}{|c|c|c|c|c|c|c|}
         \hline
         \textbf{Parameter} & \textbf{\#1} & \textbf{\#2} & \textbf{\#3} & \textbf{\#4} & \textbf{\#5} & \textbf{\#6} \\
         \hline
         \textbf{lat\_blend\_factor ($\alpha$)} & 0.0 & 0.5 & 1.0 & 0.0 & 0.5 & 1.0 \\
         \hline
         \textbf{acceleration\_exp ($\delta_{idm}$)} & \multicolumn{6}{|c|}{3.0} \\
         \hline
         \textbf{deceleration\_factor} & \multicolumn{6}{|c|}{2.0} \\
         \hline
         \textbf{max\_deceleration ($b_f$)} & \multicolumn{6}{|c|}{-4.0} \\
         \hline
         \textbf{max\_obs\_deceleration ($b_l$)} & \multicolumn{6}{|c|}{-8.0} \\
         \hline
         \textbf{path\_width\_m ($w(s)$)} & \multicolumn{6}{|c|}{12.0} \\
         \hline
         \textbf{preferred\_acceleration ($a$)} & \multicolumn{6}{|c|}{1.0} \\
         \hline
         \textbf{preferred\_deceleration ($b$)} & \multicolumn{6}{|c|}{-1.0} \\
         \hline
         \textbf{preferred\_speed ($v$)} & \multicolumn{3}{|c|}{3.0} & \multicolumn{3}{|c|}{1.0} \\
         \hline
         \textbf{preferred\_stop\_dist ($s_0$)} & \multicolumn{6}{|c|}{0.0} \\
         \hline
         \textbf{reaction\_time ($\tau$)} & \multicolumn{6}{|c|}{1.0} \\
         \hline
    \end{tabular}
    \caption{Parameters used for the command during the real tests}
    \label{tab:real_parameters}
\end{table}

\section{Presentation and Discussion of Results}
\label{sec:results}

\begin{figure}
    \centering
    \includegraphics[width=0.99\linewidth,trim=1cm 1cm 1cm 1cm]{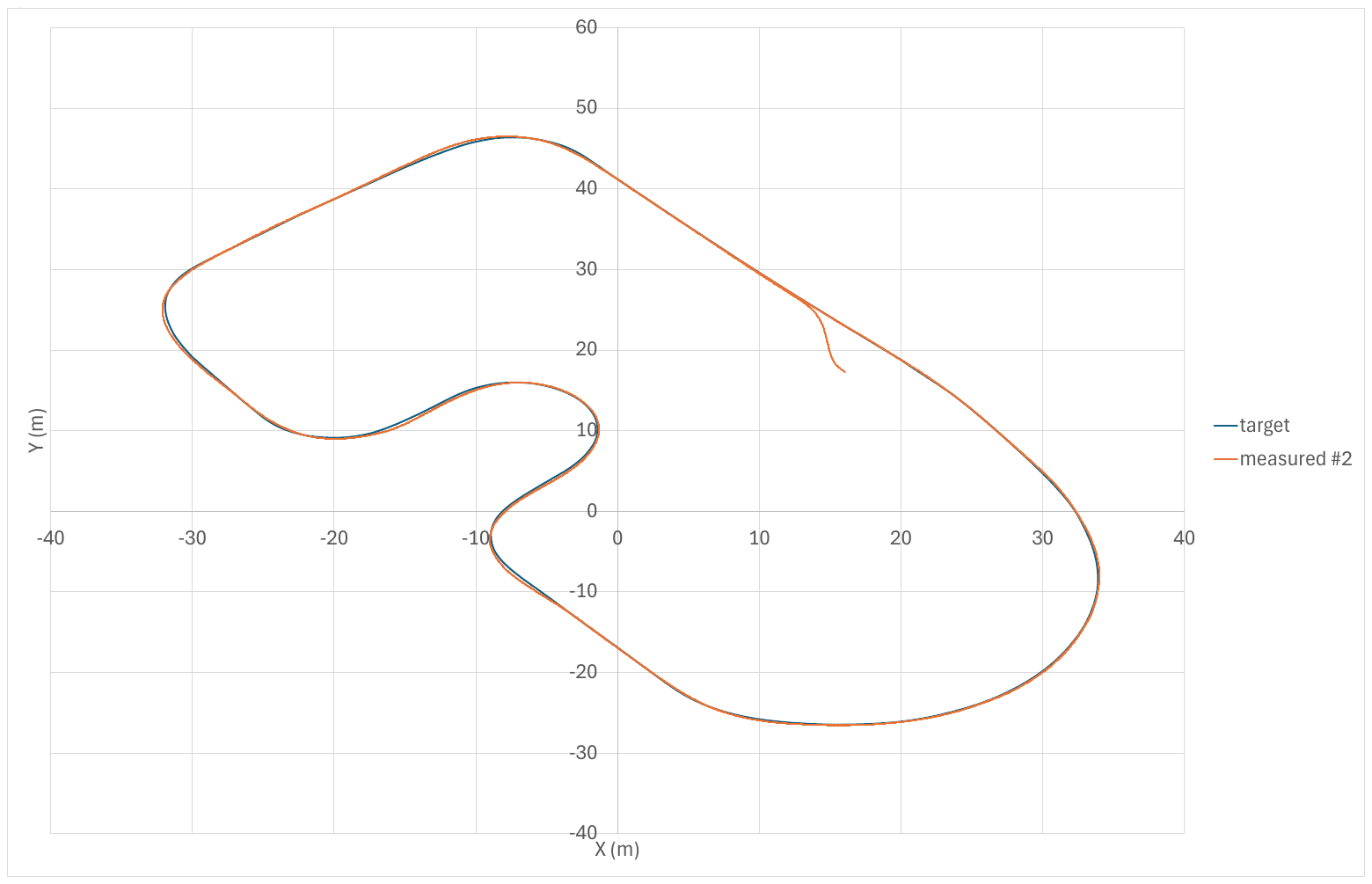}
    \caption{Measured position of the vehicle in ENU local frame during test \#2, compared to target trajectory}
    \label{fig:track}
\end{figure}

The Fig. \ref{fig:track} shows the coordinates of the track and the measured coordinates of the vehicle when it is trying to follow the track (in an ENU local frame, for test \#2), where it can be seen how close is the vehicle from the reference track.

\begin{figure}
    \centering
    \includegraphics[width=0.99\linewidth,trim=1cm 3cm 1cm 3cm]{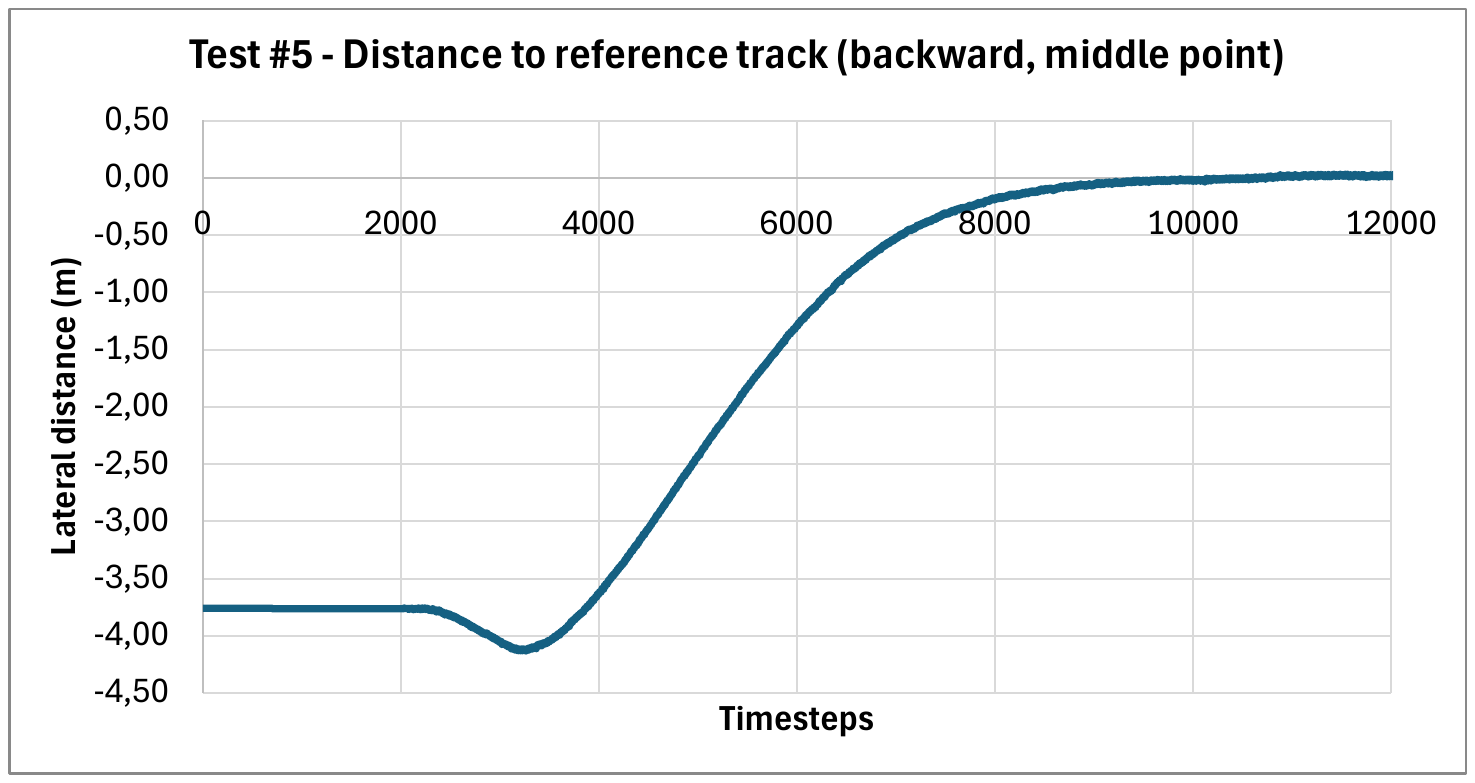}
    \caption{Lateral distance (m) between the vehicle and the reference track for test \#5 reaching the target trajectory}
    \label{fig:test_lat_distance_5_init}
\end{figure}

\begin{figure}
    \centering
    \includegraphics[width=0.99\linewidth,trim=1cm 3cm 1cm 3cm]{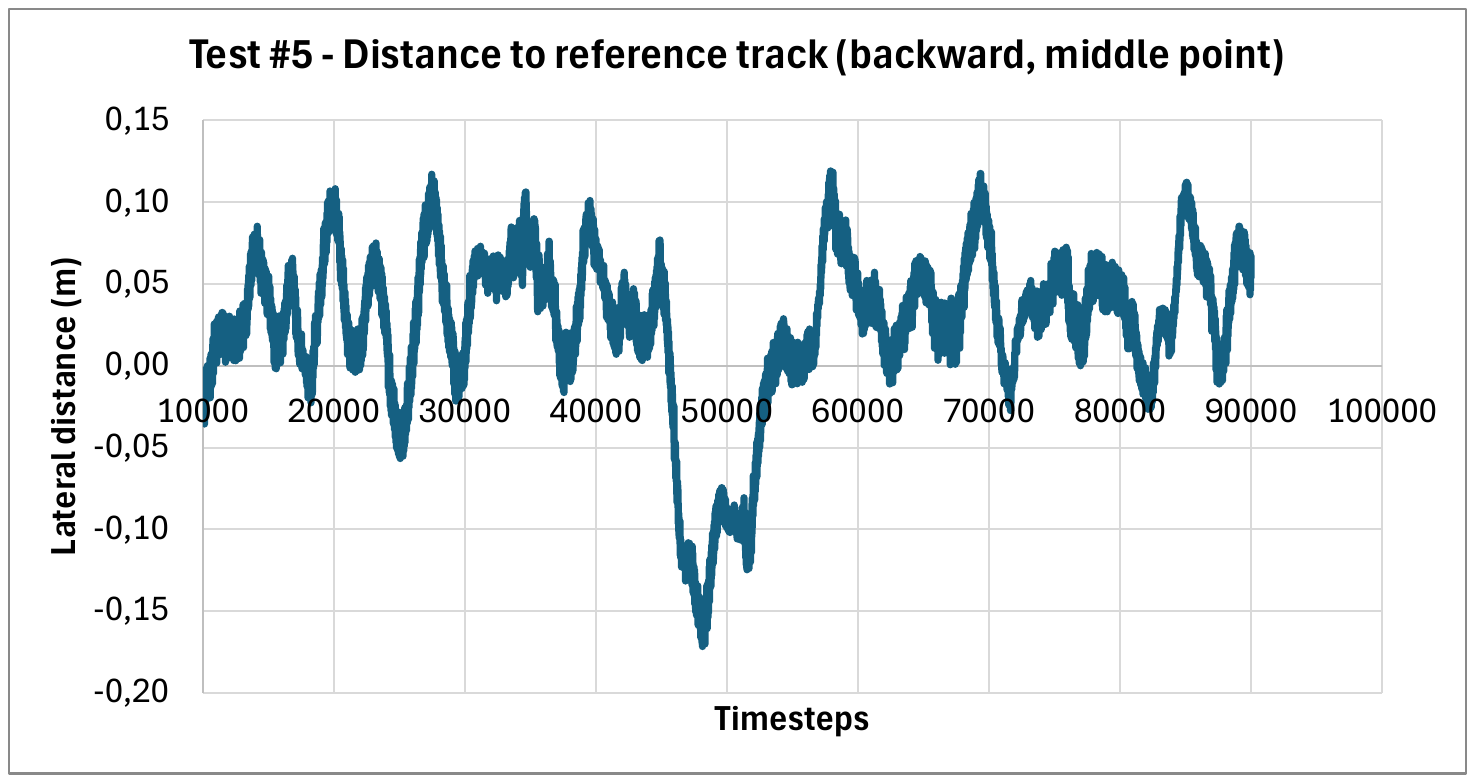}
    \caption{Lateral distance (m) between the vehicle and the reference track for test \#5 after reaching the target trajectory}
    \label{fig:test_lat_distance_5_run}
\end{figure}

\begin{figure}
    \centering
    \includegraphics[width=0.99\linewidth,trim=1cm 3cm 1cm 3cm]{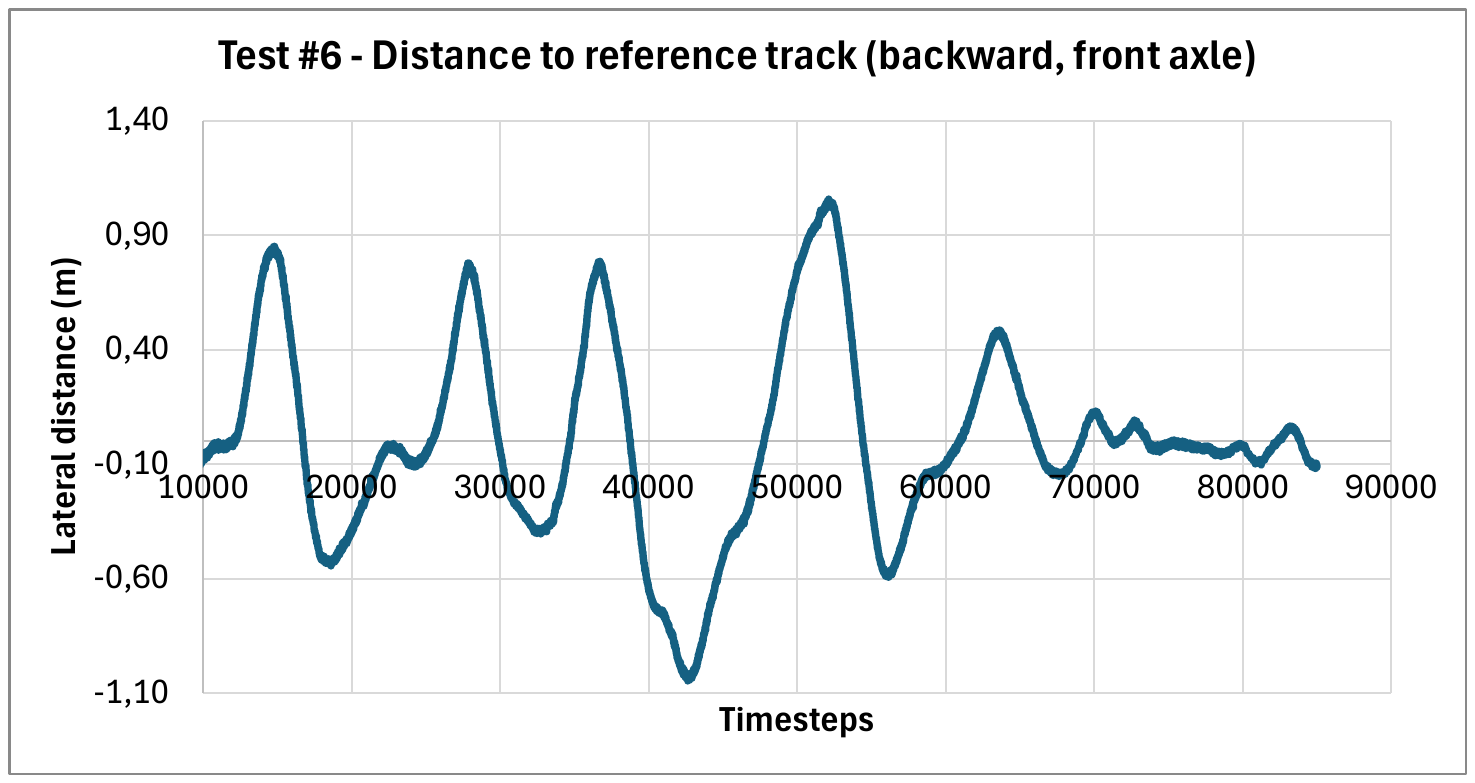}
    \caption{Lateral distance (m) between the vehicle and the reference track for test \#6 after reaching the target trajectory}
    \label{fig:test_lat_distance_6_run}
\end{figure}

In Fig. \ref{fig:test_lat_distance_5_init}, \ref{fig:test_lat_distance_5_run}, the relative distance of the control point to the track is displayed for the test \#5 (when the vehicle is joining the trajectory for Fig. \ref{fig:test_lat_distance_5_init}, and when the vehicle is following the trajectory for Fig. \ref{fig:test_lat_distance_5_run}), showing the stability and accuracy of selecting a control point in the middle.

It can be seen that test \#5 have a good accuracy (with an error always below 0.15m after the initialization), while \#6 (Fig. \ref{fig:test_lat_distance_6_run}) shows worse results. It can be explained by the fact that setting the control point on the front axle utilizes only the Stanley controller whose behavior is less accurate than the one of the command from Eq. \eqref{eq:newcf_backward}. In this case, the Fig. \ref{fig:test_lat_distance_6_run} can be seen as a baseline for backward maneuvers. On the other hand, the test \#5 with the application in the middle shows that both commands can be efficiently merged. The absolute mean distance for each test are given in Table \ref{tab:lat_errors} showing 81.2\% of distance reduction between test \#5 and \#6. The measured values also assess the stability of the command.

The measures for all tests are given in the Appendix.

\begin{table}[]
    \centering
    \begin{tabular}{|c|c|}
        \hline
         \textbf{Test number} & \textbf{Average distance (m)} \\
         \hline
         1 & 0.043\\
         \hline
         2 & 0.042\\
         \hline
         3 & 0.026\\
         \hline
         4 & 0.050\\
         \hline
         5 & 0.049 \\
         \hline
         6 & 0.260 \\
         \hline
    \end{tabular}
    \caption{Average distance between the control point and the reference track after initialization}
    \label{tab:lat_errors}
\end{table}

\begin{figure}
    \centering
    \includegraphics[width=0.99\linewidth,trim=1cm 1cm 1cm 1cm]{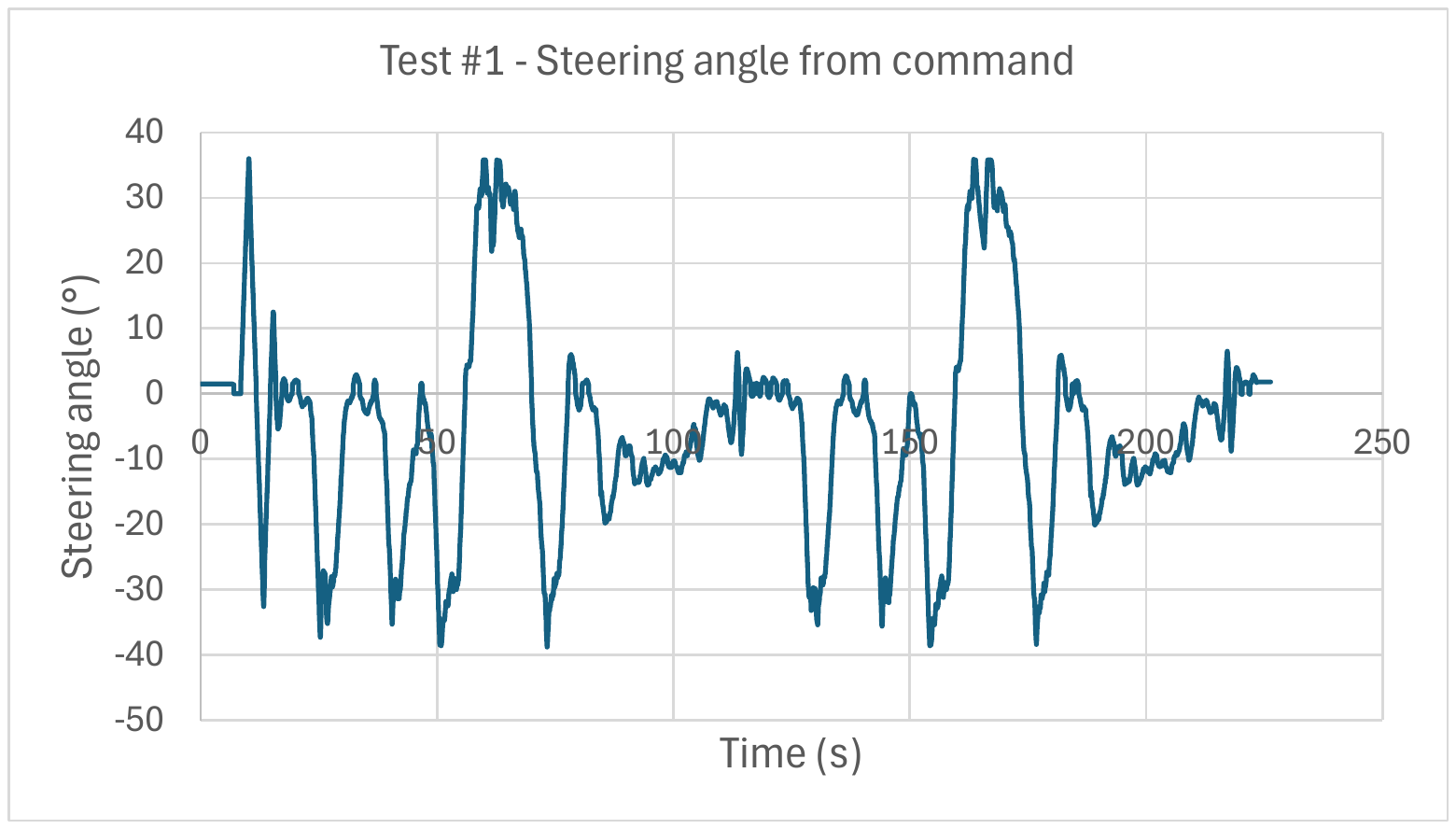}
    \caption{Steering angle (°) computed for test \#1}
    \label{fig:test1_steering_angle}
\end{figure}

\begin{figure}
    \centering
    \includegraphics[width=0.99\linewidth,trim=1cm 1cm 1cm 1cm]{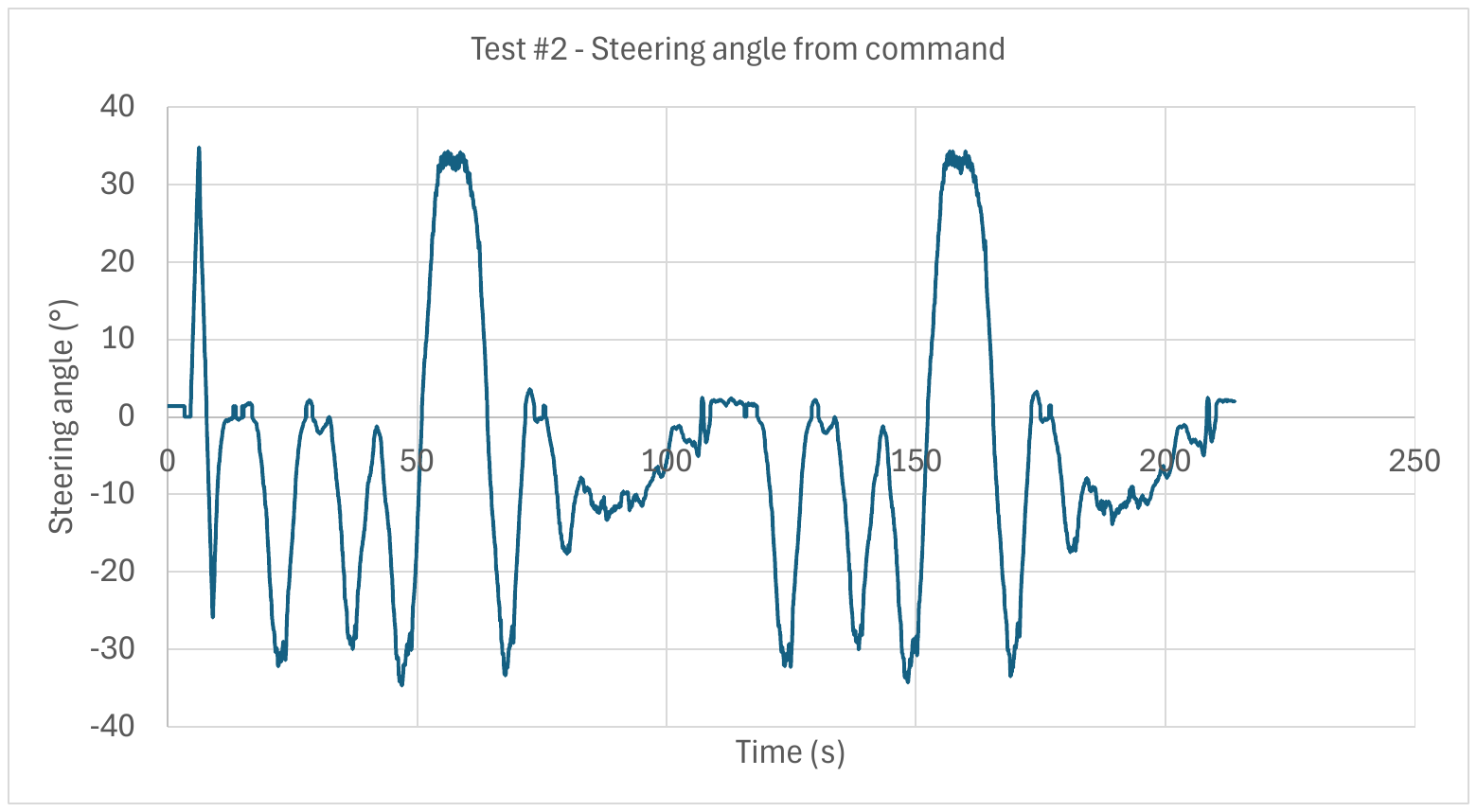}
    \caption{Steering angle (°) computed for test \#2}
    \label{fig:test2_steering_angle}
\end{figure}

\begin{figure}
    \centering
    \includegraphics[width=0.99\linewidth,trim=1cm 1cm 1cm 1cm]{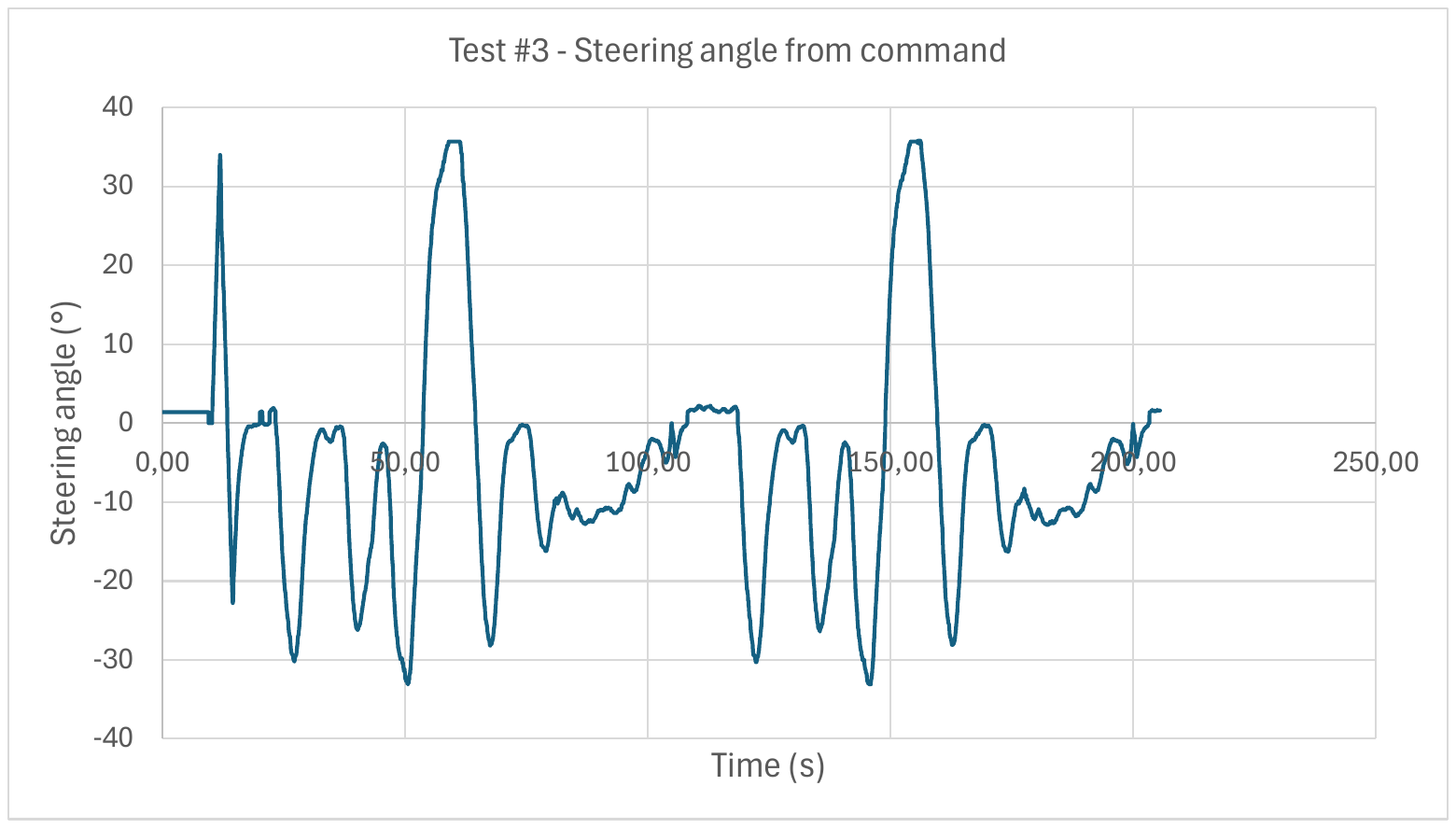}
    \caption{Steering angle (°) computed for test \#3}
    \label{fig:test3_steering_angle}
\end{figure}

\begin{figure}
    \centering
    \includegraphics[width=0.99\linewidth,trim=1cm 1cm 1cm 1cm]{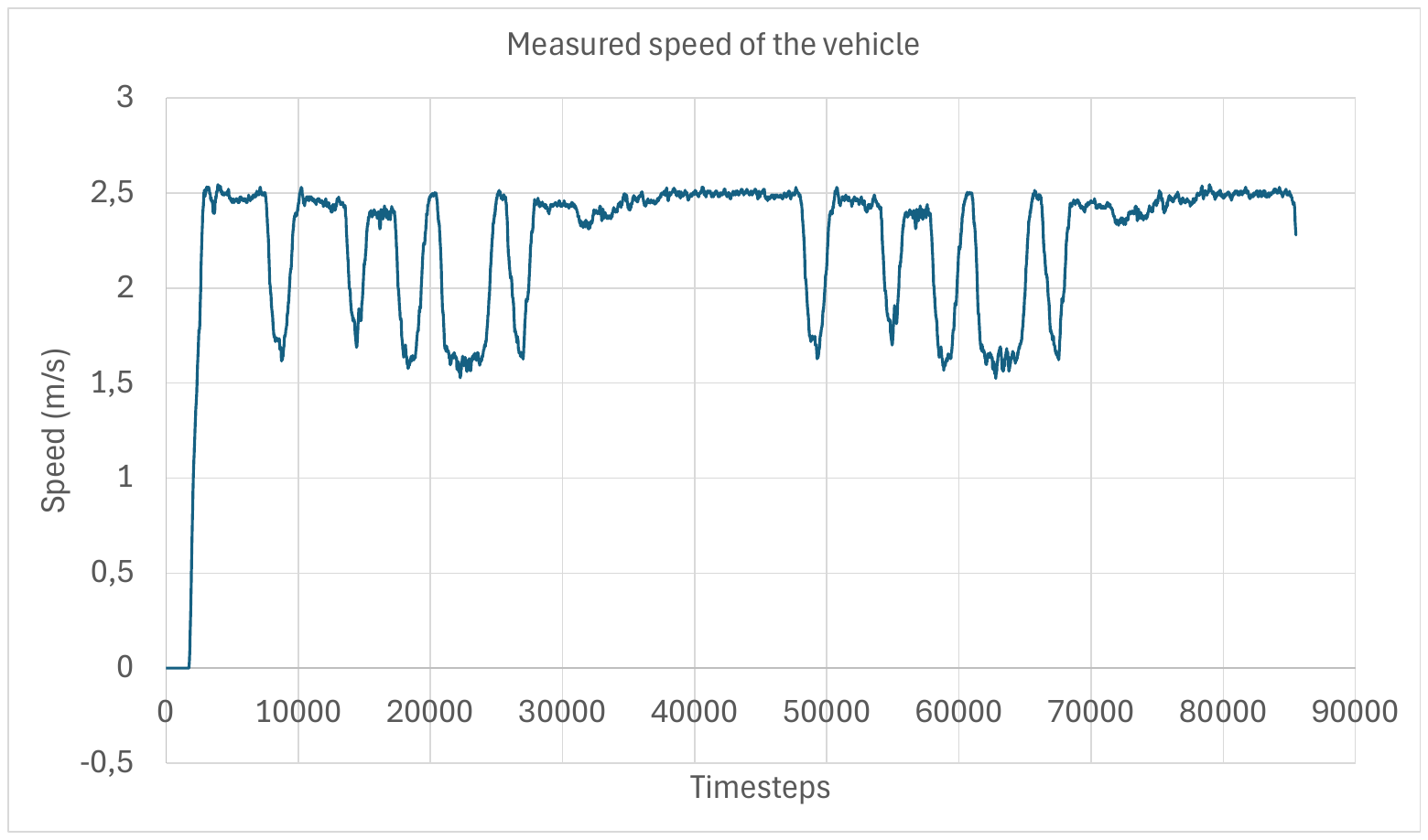}
    \caption{Speed of the vehicle (m/s) for test \#2}
    \label{fig:speed_test_2}
\end{figure}

The Fig. \ref{fig:speed_test_2} shows the speed of the vehicle during the test \#2 (the values are comparable for tests \#1 and \#3 as the longitudinal control is not directly impacted by the control point). The adaptation of the velocity along the track can be clearly seen, with a deceleration when the curvature is changing and an acceleration when it is constant. The difference between the observed target speed ($2.5m.s^{-1}$) and the desired target speed ($3.0m.s^{-1}$) is explained by the proportional controller mapping the target acceleration from the command to the throttle control.

\section{Conclusion}
\label{sec:conclusion}



This paper introduced an adaptive path-tracking framework for autonomous vehicles based on the continuous blending of lateral control commands applied at the front and rear axles, together with a curvature-aware longitudinal control strategy. By formulating the lateral steering command as a barycentric combination of a front-axle Stanley controller and a rear-axle curvature-based geometric controller, the proposed method enables a dynamic selection of the effective control point along the wheelbase. This flexibility allows the vehicle to seamlessly adapt its behavior across a wide range of driving contexts, including low-speed maneuvers, tight turns, and backward motion.

Experimental results obtained both in simulation and on a real autonomous vehicle demonstrate that the proposed approach significantly improves trajectory tracking accuracy and steering smoothness compared to fixed control-point baselines. In particular, the intermediate control-point configuration consistently achieved the best compromise between accuracy and stability, with measured lateral errors reduced by more than 80\% in backward maneuvers relative to a front-axle-only strategy. The experiments also confirm that the blending formulation remains stable in the presence of sensor noise and actuation delays, validating its practical applicability.

In addition, the proposed virtual-border-based longitudinal control effectively anticipates upcoming curvature variations and regulates vehicle speed accordingly. By converting geometric constraints into a virtual obstacle distance through ray tracing, the method provides a simple yet robust mechanism to adapt speed without requiring explicit curvature thresholds or complex optimization. The resulting speed profiles exhibit smooth deceleration and acceleration phases that contribute to improved lateral tracking performance and overall vehicle safety.

Overall, the proposed control architecture combines simplicity, robustness, and adaptability, leveraging well-established control laws while extending their operational domain through dynamic control-point blending. Future work will focus on formal stability analysis of the blended control law, automatic adaptation of the blending factor based on vehicle state and environment perception, and extension to the extended bicycle model, to handle tractor-trailer systems. The integration of the proposed framework within predictive or learning-based supervisory layers also constitutes a promising direction for further research.

\section{Bibliography}
\label{sec:bibliography}

\bibliographystyle{IEEEtran}
\bibliography{biblio.bib}

\section{Appendices}

\subsection{Appendix: Signed distance for tests}
\label{appendix:results}

\begin{figure}[h!]
    \centering
    \includegraphics[width=0.99\linewidth, trim=1cm 3cm 1cm 3cm]{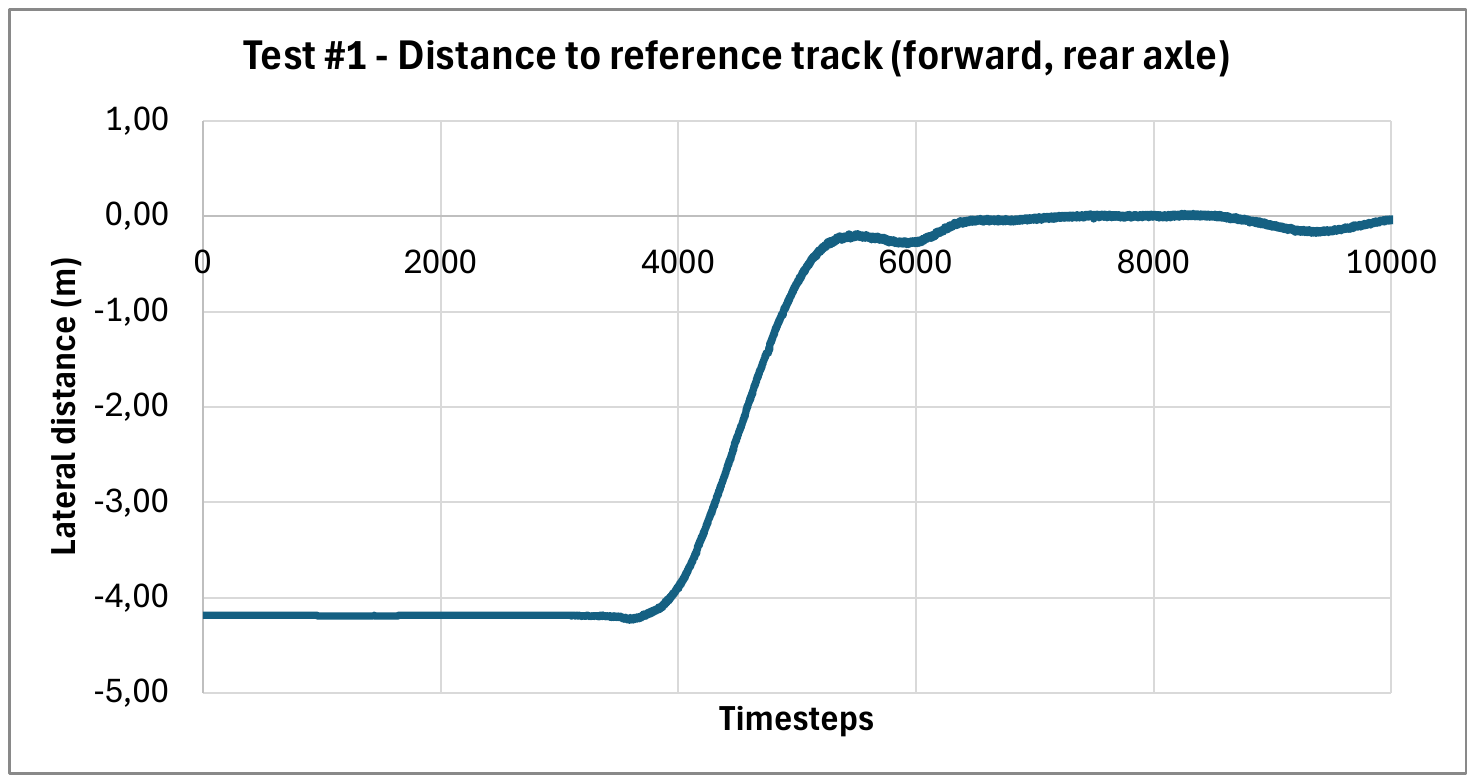}
    \caption{Lateral distance for test \#1 when reaching the target trajectory}
    \label{fig:placeholder}
\end{figure}

\begin{figure}[h!]
    \centering
    \includegraphics[width=0.99\linewidth, trim=1cm 3cm 1cm 3cm]{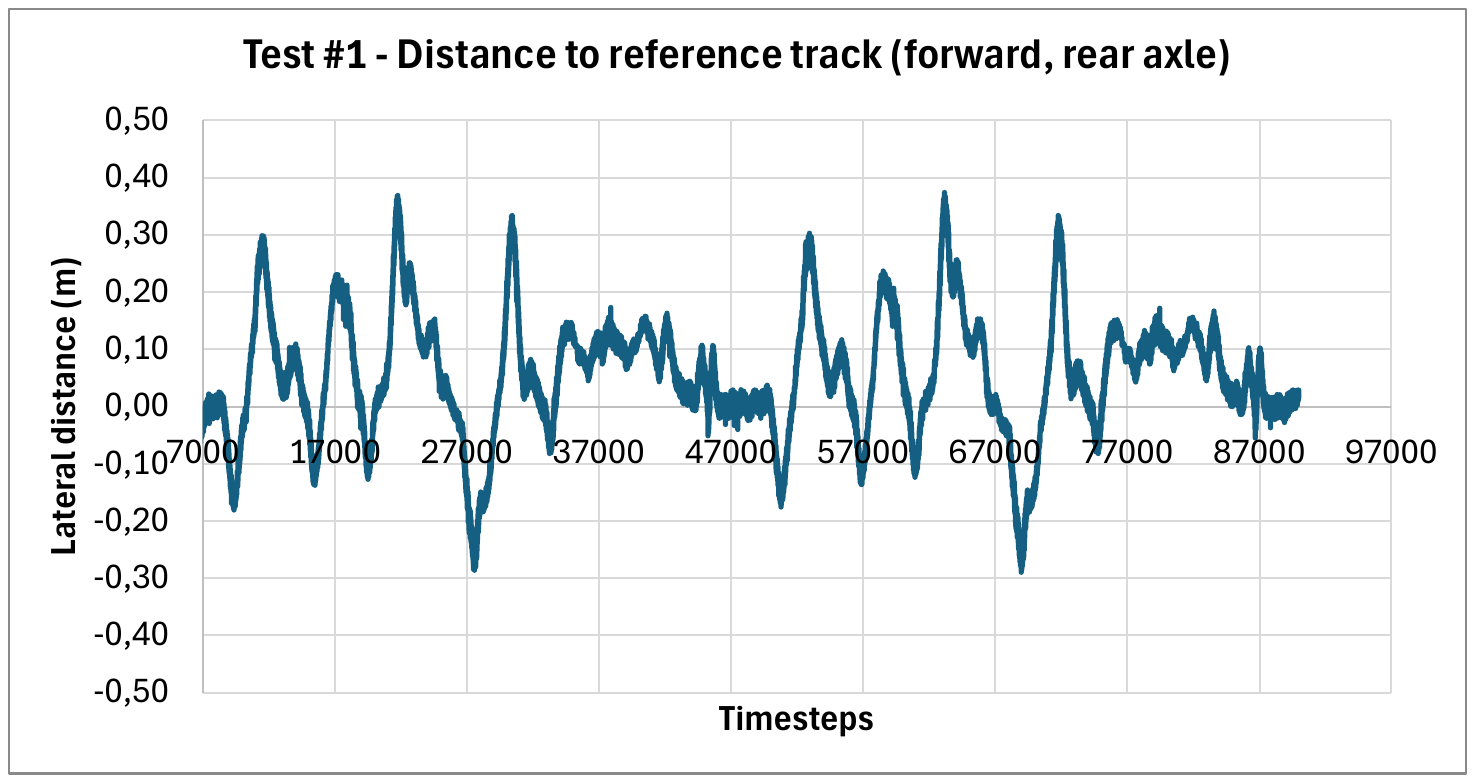}
    \caption{Lateral distance for test \#1 after reaching the target trajectory}
    \label{fig:placeholder}
\end{figure}

\begin{figure}[h!]
    \centering
    \includegraphics[width=0.99\linewidth, trim=1cm 3cm 1cm 3cm]{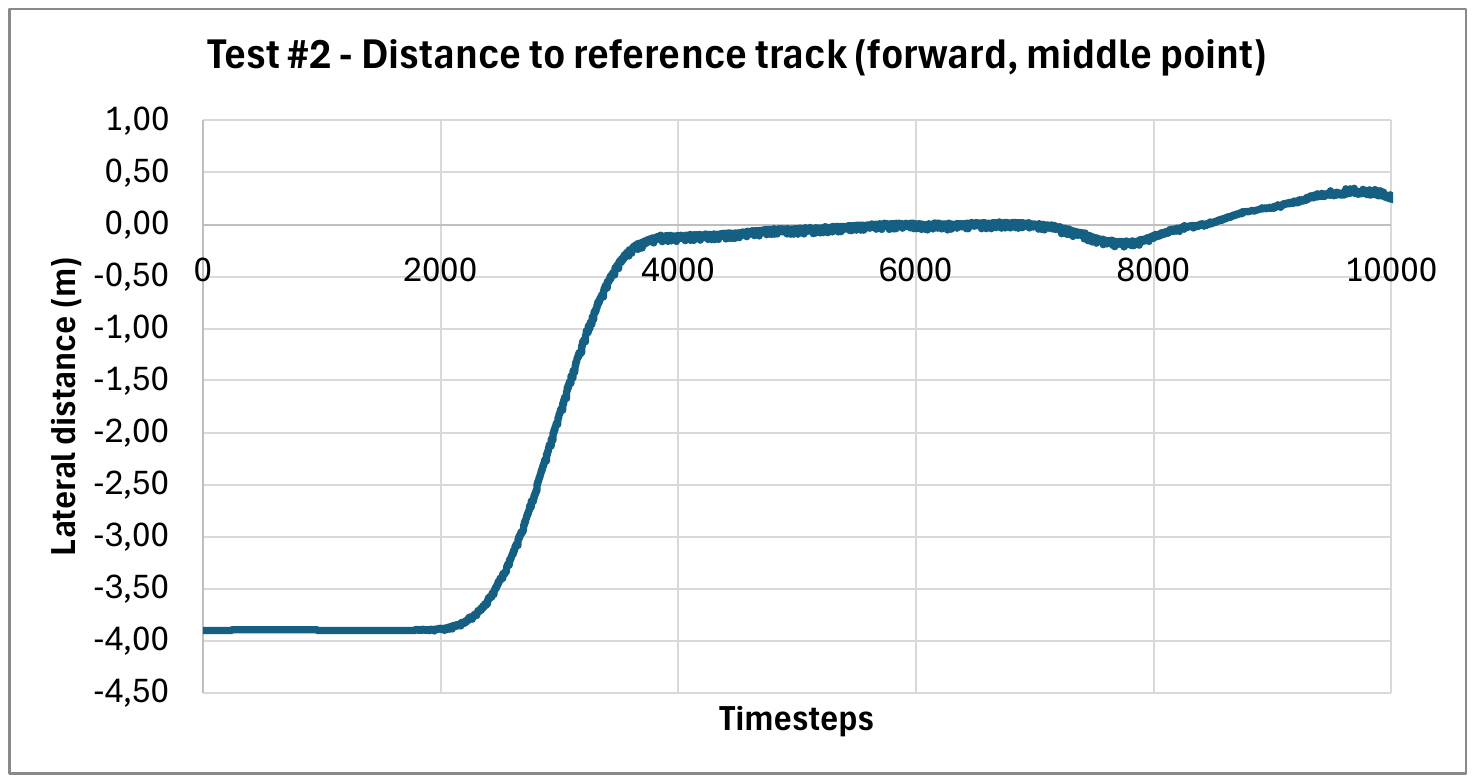}
    \caption{Lateral distance for test \#2 when reaching the target trajectory}
    \label{fig:placeholder}
\end{figure}

\begin{figure}[h!]
    \centering
    \includegraphics[width=0.99\linewidth, trim=1cm 3cm 1cm 3cm]{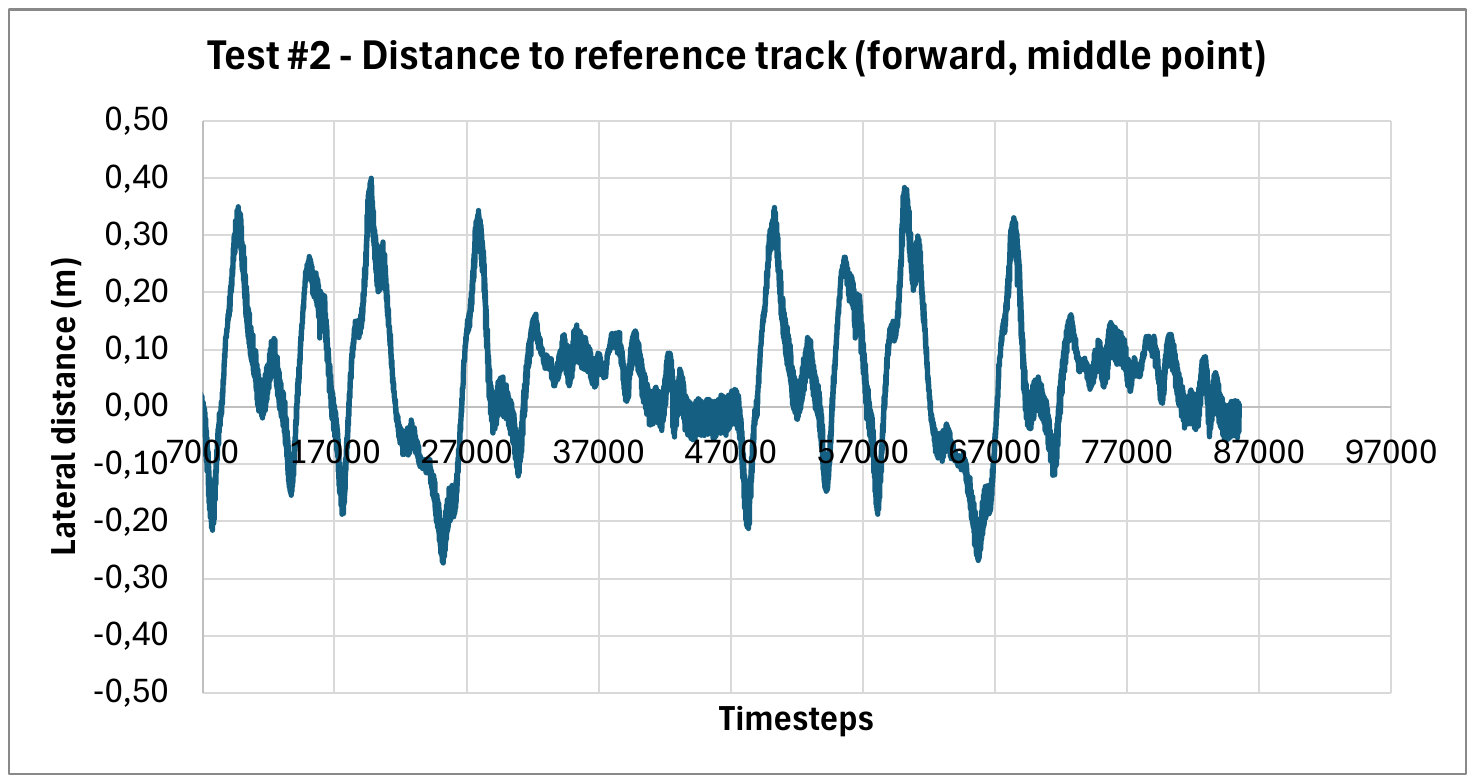}
    \caption{Lateral distance for test \#2 after reaching the target trajectory}
    \label{fig:placeholder}
\end{figure}

\begin{figure}[h!]
    \centering
    \includegraphics[width=0.99\linewidth, trim=1cm 3cm 1cm 3cm]{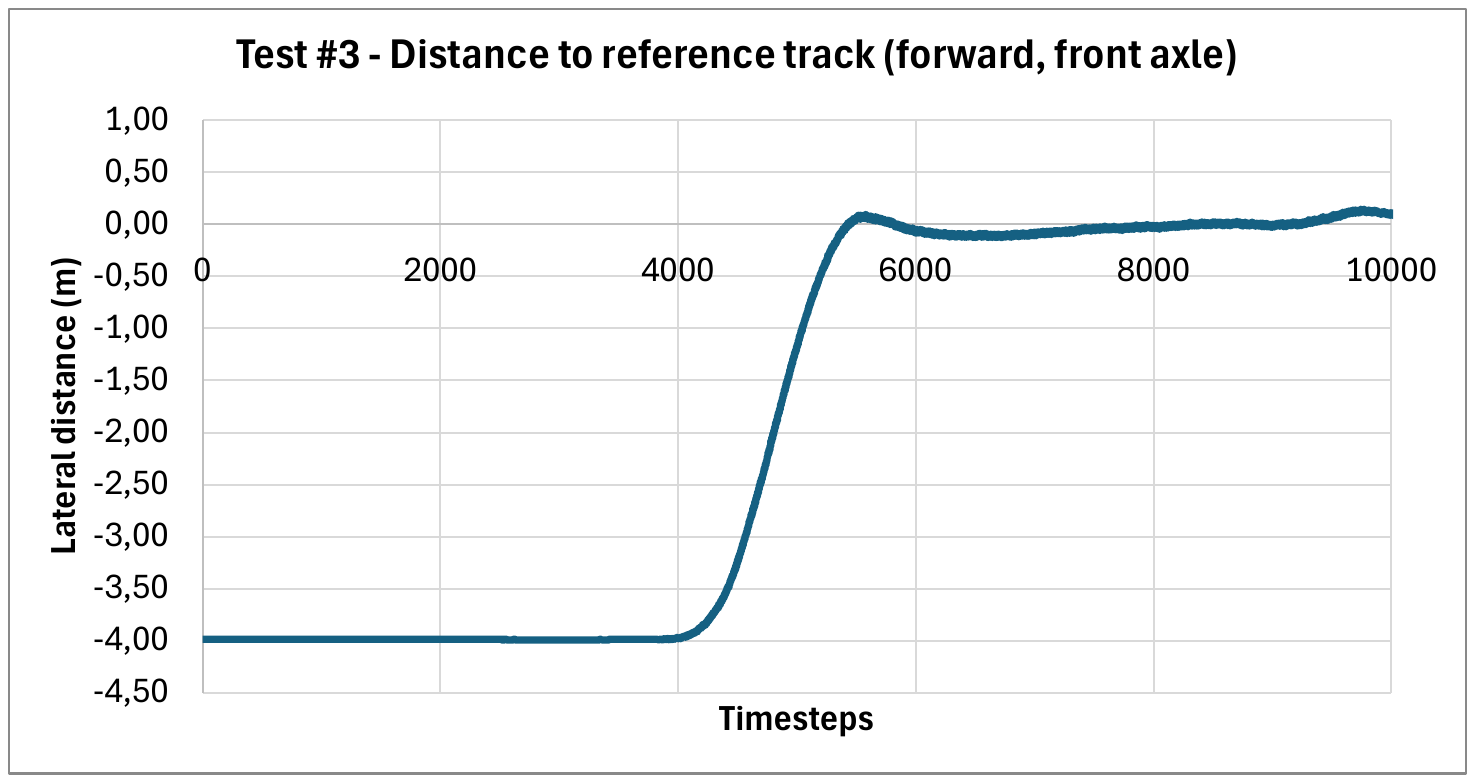}
    \caption{Lateral distance for test \#3 when reaching the target trajectory}
    \label{fig:placeholder}
\end{figure}

\begin{figure}[h!]
    \centering
    \includegraphics[width=0.99\linewidth, trim=1cm 3cm 1cm 3cm]{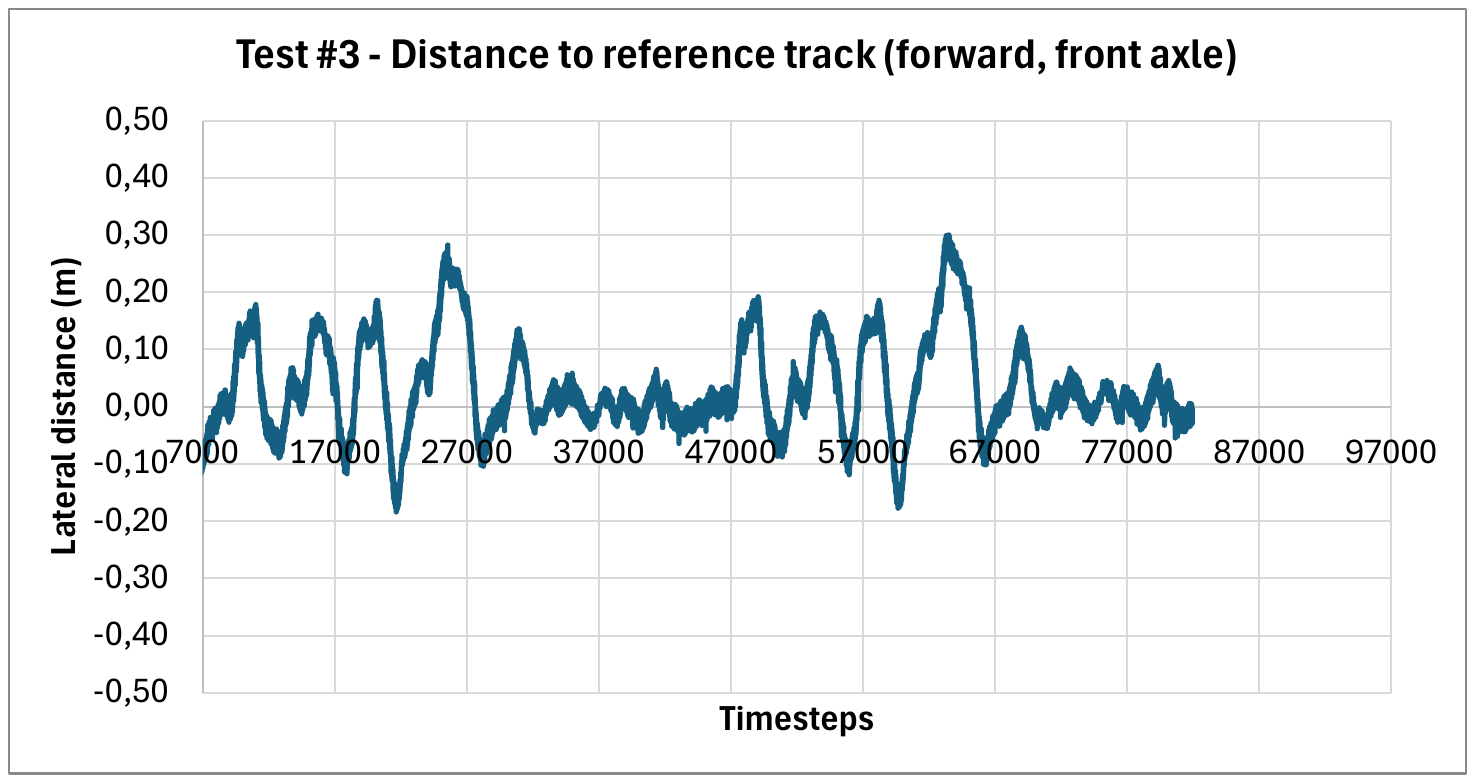}
    \caption{Lateral distance for test \#3 after reaching the target trajectory}
    \label{fig:placeholder}
\end{figure}

\begin{figure}[h!]
    \centering
    \includegraphics[width=0.99\linewidth, trim=1cm 3cm 1cm 3cm]{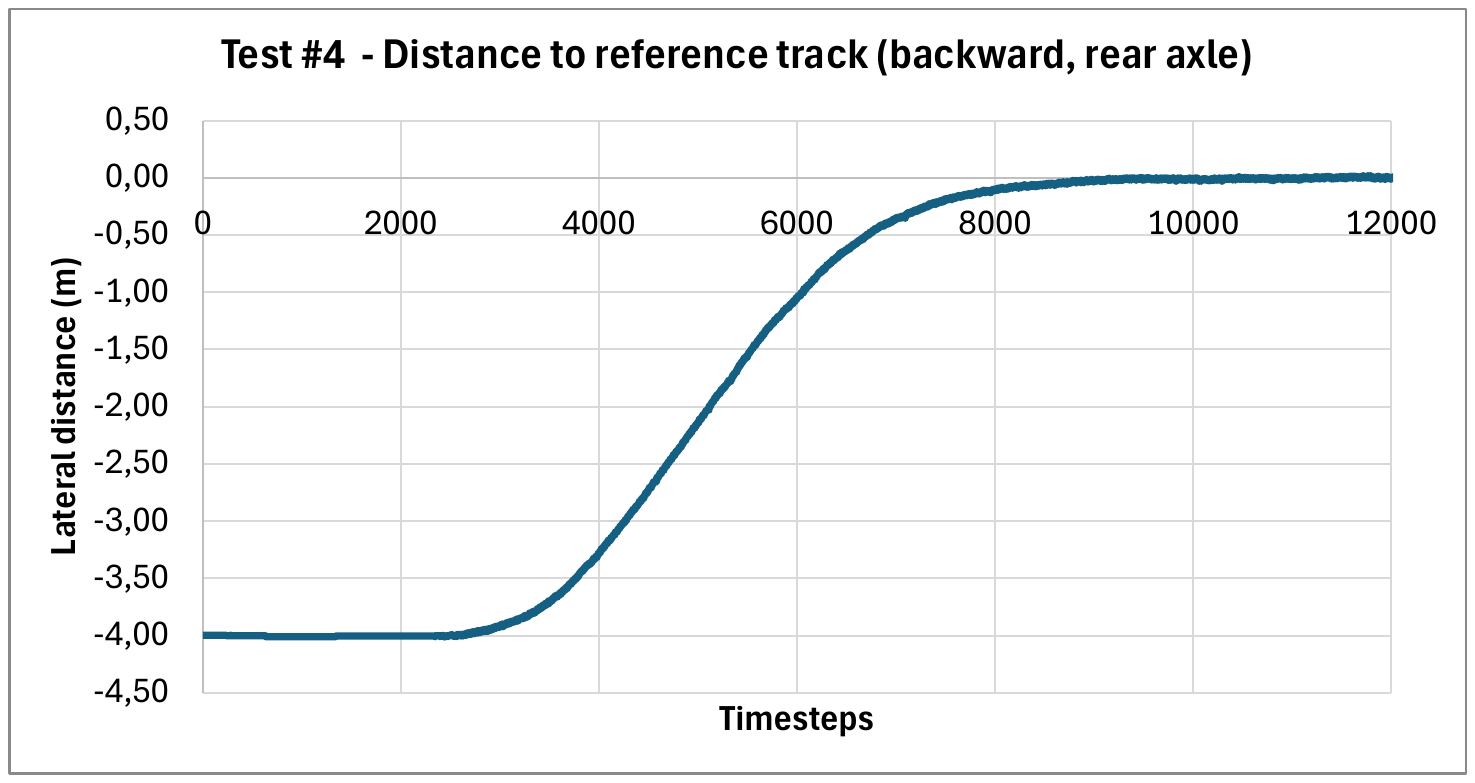}
    \caption{Lateral distance for test \#4 when reaching the target trajectory}
    \label{fig:placeholder}
\end{figure}

\begin{figure}[h!]
    \centering
    \includegraphics[width=0.99\linewidth, trim=1cm 3cm 1cm 3cm]{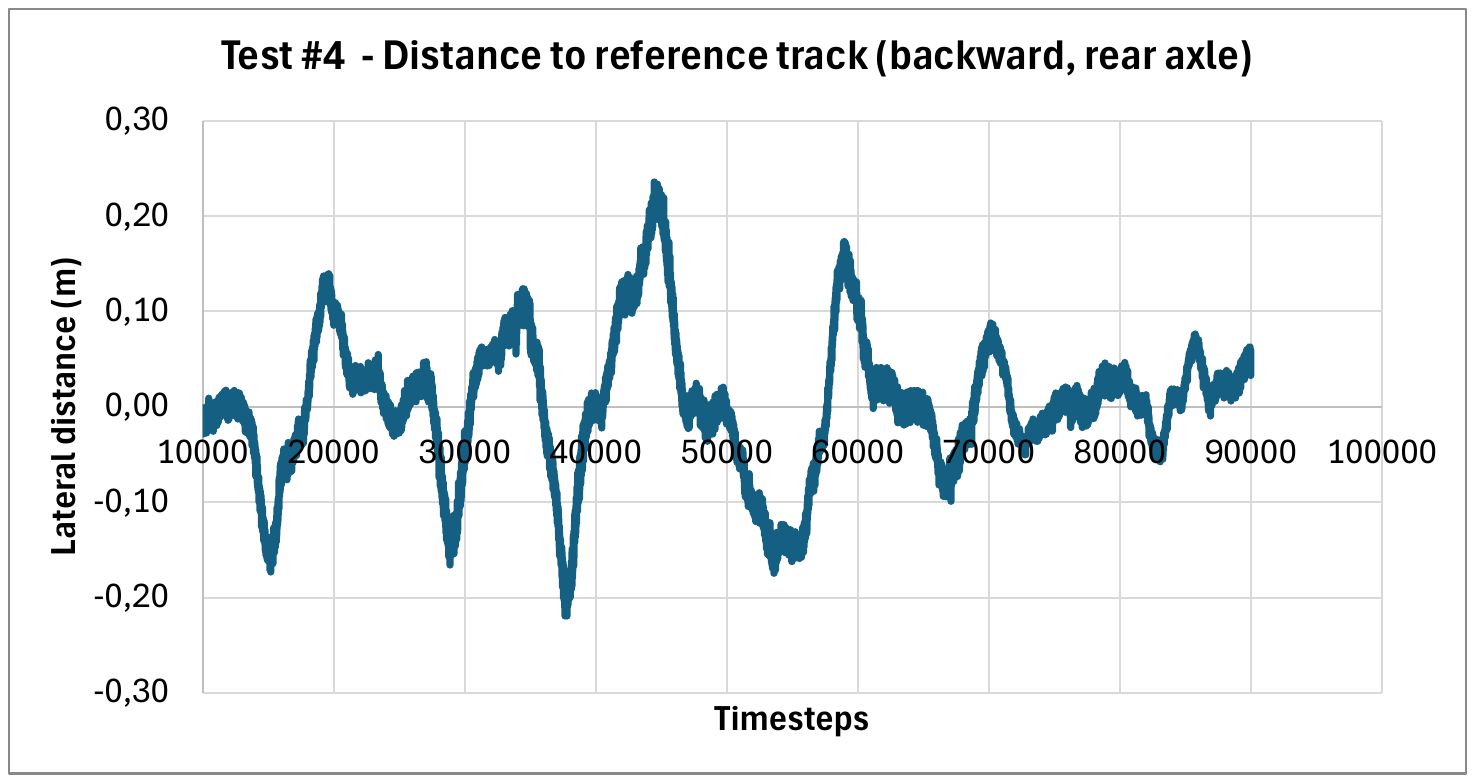}
    \caption{Lateral distance for test \#4 after reaching the target trajectory}
    \label{fig:placeholder}
\end{figure}

\begin{figure}[h!]
    \centering
    \includegraphics[width=0.99\linewidth, trim=1cm 3cm 1cm 3cm]{imgs/signed_distance/test5_init.pdf}
    \caption{Lateral distance for test \#5 when reaching the target trajectory}
    \label{fig:placeholder}
\end{figure}

\begin{figure}[h!]
    \centering
    \includegraphics[width=0.99\linewidth, trim=1cm 3cm 1cm 3cm]{imgs/signed_distance/test5_run.pdf}
    \caption{Lateral distance for test \#5 after reaching the target trajectory}
    \label{fig:placeholder}
\end{figure}

\begin{figure}[h!]
    \centering
    \includegraphics[width=0.99\linewidth, trim=1cm 3cm 1cm 3cm]{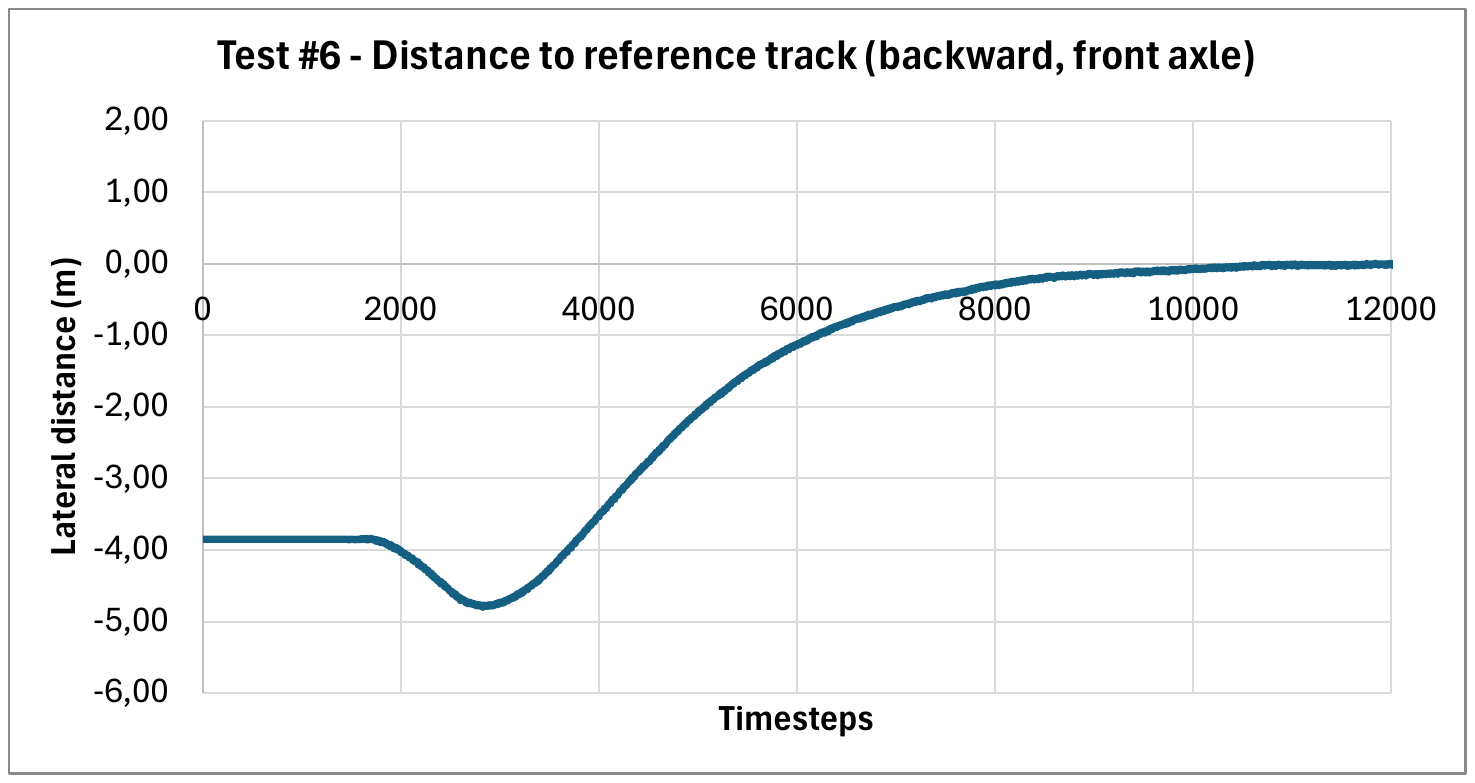}
    \caption{Lateral distance for test \#6 when reaching the target trajectory}
    \label{fig:placeholder}
\end{figure}

\begin{figure}[h!]
    \centering
    \includegraphics[width=0.99\linewidth, trim=1cm 3cm 1cm 3cm]{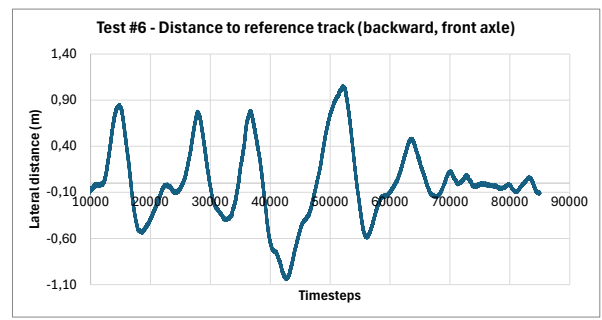}
    \caption{Lateral distance for test \#6 after reaching the target trajectory}
    \label{fig:placeholder}
\end{figure}

\end{document}